\documentclass[preprint,12pt,numbers]{elsarticle}

\usepackage{amssymb}
\usepackage{url}
\usepackage{multirow}
\usepackage{multirow}
\usepackage{makecell}
\usepackage{caption}
\usepackage{subfig}
\usepackage{soul}
\usepackage{color}
\usepackage{lipsum}
\usepackage{amsmath}
\usepackage{hyperref}
\usepackage{graphicx}
\usepackage{mathtools}
\usepackage{multirow}
\usepackage[marginal]{footmisc}
\usepackage{ntheorem}
\usepackage{amssymb}
\usepackage{algorithmic}
\usepackage{capt-of}
\usepackage{url}

\newtheorem{Lemma}{Lemma}

\usepackage{graphicx}

\usepackage{multirow}
\usepackage{newtxtext,newtxmath}
\usepackage{graphicx}
\usepackage{natbib}
\usepackage{algorithm}
\usepackage[algo2e]{algorithm2e}
\usepackage{algorithmic}
\usepackage{setspace}
\usepackage{marvosym}
\usepackage{hyperref}
\usepackage[numbers]{natbib}
\journal{Pattern Recognition}

\begin{document}

\begin{frontmatter}



\title{Learning Node Representations against Perturbations}

\author[label1,label2]{Xu Chen}
\author[label3]{Yuangang Pan}
\author[label2,label3]{Ivor Tsang}
\author[label1]{Ya Zhang\textsuperscript{\Letter}}
\address[label1]{Shanghai Jiao Tong University,~Shanghai,~China}
\address[label2]{University of Technology Sydney,~Sydney,~Australia}
\address[label3]{Center for Frontier AI Research Research, Agency for Science Technology, Singapore}

% \author[label1,label2]{Xu Chen}
% \author[label2]{Yuangang Pan}
% \author[label2]{Ivor Tsang}
% \author[label1]{Ya Zhang*}

% \affiliation[label1]{
%             organization={Shanghai Jiao Tong University},
%             % addressline={}, 
%             city={Shanghai},
%             % postcode={}, 
%             % state={},
%             country={China}
%             }
% \affiliation[label2]{
%             organization={University of Technology Sydney},
%             % addressline={}, 
%             city={Sydney},
%             % postcode={}, 
%             % state={},
%             country={Australia}
%             }

\begin{abstract}
Recent graph neural networks (GNN) has achieved remarkable performance in node representation learning\footnote{xuchen2016@sjtu.edu.cn;pan\_yuangang@ihpc.a-star.edu.sg}\footnote{ivor.tsang@uts.edu.au;ya\_zhang@sjtu.edu.cn}. One key factor of GNN's success is the \emph{smoothness} property on node representations. Despite this, most GNN models are fragile to the perturbations on graph inputs and could learn unreliable node representations.
In this paper, we study how to learn node representations against perturbations in GNN.
Specifically, we consider that a node representation should remain stable under slight perturbations on the input, and node representations from different structures should be identifiable,
which two are termed as the \emph{stability} and \emph{identifiability} on node representations, respectively. To this end, we propose a novel model called  Stability-Identifiability GNN Against Perturbations (SIGNNAP) that learns reliable node representations in an unsupervised manner. SIGNNAP formalizes the \emph{stability} and \emph{identifiability} by a contrastive objective and preserves the \emph{smoothness} with existing GNN backbones. The proposed method is a generic framework that can be equipped with many other backbone models (e.g. GCN, GraphSage and GAT). Extensive experiments on six benchmarks under both transductive and inductive learning setups of node classification demonstrate the effectiveness of our method. Codes and data are available online:~\url{https://github.com/xuChenSJTU/SIGNNAP-master-online}
\end{abstract}

\end{frontmatter}


\section{Introduction}
Learning node representations for graphs is an important research topic that has great promise in a variety of areas~\cite{YAO2022108708,HAO2023109504}.
In recent years, it has been studied by two main methods: the network embedding based methods and the graph neural network (GNN) based methods.
% ~\cite{wu2019comprehensive}. 
The network embedding based methods such as DeepWalk~\cite{perozzi2014deepwalk} and Node2Vec~\cite{grover2016node2vec} mainly use the statistical random walks on graphs by a language model. DeepWalk and Node2Vec benefit the learning of node representations on graphs while rely on high-quality random walks. On the other hand, inspired by the graph convolutional theory and deep learning, GNN has emerged as one crucial technique for node representation learning on graphs.
\textcolor{black}{The concept of GNN was firstly proposed in~\cite{1555942}. Later, ChebNet~\cite{defferrard2016convolutional}, introduces a fast localized convolution approach on graphs in spectral domain.} Inspired by the idea that high-order convolutions can be built by stacking multiple convolutional layers,
graph convolutional network (GCN)~\cite{kipf2016semi} simplifies ChebNet with multiple stacked graph convolutional layers where each layer is one-hop convolution. GCN also connects the spectral graph convolution with information propagation, which encourages the following design of GNN models~\cite{velickovic2017graph,hamilton2017inductive}.

\begin{figure*}[t]
\centering
\includegraphics[width=12.0cm]{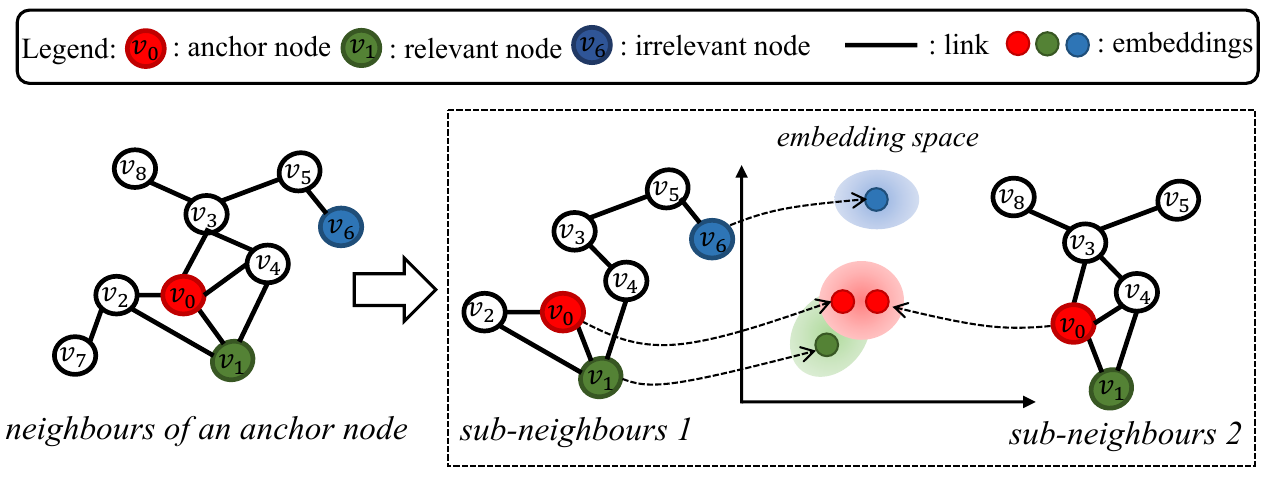}
% \vspace{-5pt}
\caption{An example to illustrate the
\emph{smoothness}, \emph{stability} and \emph{identifiability} of node representations on graphs. For an anchor node $v_{0}$, let $v_{1}$ be a relevant node and $v_{6}$ be an irrelevant node. Here, different sub-neighbours refer to different variants of the original neighbours under slight perturbations (e.g. drop edge). A reliable GNN model should preserve the \emph{smoothness} (i.e. the embeddings of relevant nodes ($v_{0}$ and $v_{1}$) are close to each other), the \emph{stability} (i.e. the representations of $v_{0}$ is stable to slight perturbations) and the \emph{identifiability} (i.e. the embeddings of irrelevant nodes ($v_{0}$ and $v_{6}$) are distant from each other).}
\label{figure:motivation}
\vspace{-10pt}
\end{figure*}

A key characteristic of GNN is smoothing which aggregates the features of a node and its nearby neighbours~\cite{li2018deeper}, enforcing \emph{smoothness} on node representations. However, most GNN models are fragile to the perturbations on the input and learn unreliable node representations~\cite{verma2019stability,rong2019dropedge}.
Studying how to learn node representations against perturbations in GNN is a promising topic. When learning against perturbations, a node's representation should be stable to the slight perturbations on graph inputs, which can be termed as the \emph{stability} of node representations. Meanwhile, in order to emphasize the graph signals against perturbations, nodes of different structures should have identifiable representations, i.e. the \emph{identifiability} of node representations.
% Without \emph{stability}, most GNN methods would be sensitive to the slight perturbations and suffer from high variance~\cite{verma2019stability,rong2019dropedge}, leading to a non-reliable feature estimator for testing~\cite{saad2011robust}. When considering the \emph{identifiability}, we may prevent the over-smoothing issue of GNN which indicates  
In Figure~\ref{figure:motivation}, we provide an example to illustrate the \emph{smoothness}, \emph{identifiability} and \emph{stability} for node representations on graphs.
Designing a model that simultaneously enforces the three properties is important for node representation learning on graphs. However, there are several difficulties: 1) The inappropriate way of constructing perturbations may introduce bias in \emph{stability} and \emph{identifiability} estimation. 2) How to formulate the optimization objective of \emph{stability} and \emph{identifiability} determines the estimation quality.

In this paper, we propose a new model named Stability-Identifiability GNN Against Perturbations (SIGNNAP) to learn reliable node representations on graphs. Specifically, the \emph{stability} and \emph{identifiability} are interpreted as high similarity for representations of the same node with different perturbations and low similarity for representations of different nodes, respectively, which is formulated as a contrastive objective function. In SIGNNAP, we study several general perturbations on edges or node attributes with explicit distributions so that the contrastive objective can marginalize it to avoid biased \emph{stability} and \emph{identifiability} estimation.
More importantly, the proposed model is generic and can be equipped with different GNN backbones such as GCN~\cite{kipf2016semi}, GraphSage~\cite{hamilton2017inductive} and GAT~\cite{velickovic2017graph}.
By optimizing the objective in our model, we show that we can learn more reliable node representations that benefit the downstream node classification task. 
With the \emph{stability} and \emph{identifiability}, we empirically show that the model has better ability of preventing the over-smoothing and over-fitting problem on graphs. The contributions are summarized as follows:
% \vspace{-8pt}
\begin{itemize}
    \item To learn node representations against perturbations in GNN, we advocate to guarantee the \emph{stability} and \emph{identifiability} property and develop a novel model named SIGNNAP that learns reliable node representations in an unsupervised manner. The model is generic and can be equipped with many popular GNN backbone models for boosted performance.
    \vspace{-10pt}
    \item We conduct extensive experiments on six benchmarks under both transductive and inductive learning setups. 
    The results show that SIGNNAP has better ability of preventing the over-smoothing and over-fitting issue and show superior representation learning performance.
\end{itemize}
\vspace{-10pt}

\section{Related Work}
Recent progress of node representation learning on graphs is mainly categorized into two groups: network embedding and GNN~\cite{wu2019comprehensive}.

\textbf{Network Embedding}: Network embedding arises as one hot research topic to learn representative node embeddings for a given network. Various methods have been proposed for network embedding. For example, inspired by the skip-gram model for word representation in Natural Language Processing (NLP),
% ~\cite{mikolov2013distributed}, DeepWalk~\cite{perozzi2014deepwalk} learns node embeddings from random walk sequences in social networks. 
LINE~\cite{tang2015line} defines first-order and second-order proximity to describe the context of a node and trains node embeddings via negative sampling.
Node2Vec~\cite{grover2016node2vec} extends DeepWalk by designing a biased random walk to control the Bread First Search (BFS) and Deep First Search (DFS). 
% Qiu et al.~\cite{qiu2018network} analyzed DeepWalk and Node2Vec and unified them into one matrix factorization framework. 
% Other researchers extend DeepWalk with node attributes in an end-to-end learning style~\cite{chen2019}.
% Since DeepWalk and Node2Vec cannot incorporate node attributes in an end-to-end manner, then researchers consider random walks on the node-attribute bipartite graph, which is also called attributed random walk~\cite{huang2019graph,chen2019}. 
% GraphRNA~\cite{huang2019graph} takes node attributes as a bipartite graph and use it to obtain diverse random walks. Then a RNN network is used to encode nodes into low-dimensional embeddings. The attributed random walk based methods are summarized as a matrix factorization formulation in~\cite{chen2019}. 
These methods rely much on high-quality random walks and are labour-consuming because of carefully designed sampling strategies and fine-tuned hyper-parameters.

\textbf{Graph Neural Networks (GNN)}: Inspired by the success of convolution on images in Euclidean space, researchers tried to define the convolution on graphs in non-Euclidean space. In ChebNet~\cite{defferrard2016convolutional}, a fast and localized convolution filter is defined on graphs in spectral domain. In~\cite{kipf2017semi}, the authors proposed graph convolutional networks (GCN) which utilizes a localized first-order approximation of the convolution in ChebNet. 
GCN connects the spectral graph convolution to information propagation, and broadens the way of other GNN models such as GraphSage~\cite{ying2018graph} and graph attention networks (GAT)~\cite{velickovic2018graph}.  
% ARGA~\cite{pan2018adversarially} advocates to impose prior distribution on the graph embeddings by adversarial learning.
% Ying et al.~\cite{ying2018graph} defined the convolution on graphs with different neighborhood aggregation functions and proposed GraphSage which supports inductive learning on large-scale graphs. Petar et al.~\cite{velickovic2018graph} introduced attention mechanism to GCN and proposed graph attention networks (GAT).
Meanwhile, some works~\cite{li2018deeper,li2019label,ZHENG2022108492} reveal that the key of GCN's success is its smoothing characteristic that aggregates node features from the nearby neighbours. 
% Following this, GSSNN~\cite{aaai20wan} study the smoothing characteristic by heat kernel~\cite{chung1997spectral} and smoothing splines, respectively. 
Although the smoothing characteristic has put forward GNN's progress, when stacking too many layers, the GNN model usually faces the over-smoothing problem where all node representations converge to the same subspace and are unidentifiable from each other.
% ~\cite{li2019deepgcns,rong2019dropedge}.
% This problem has been analyzed by several works. For example, inspired by the residual connections in CNN, 
Accordingly, ResGCN~\cite{li2019deepgcns} introduces residual connections in GCN. 
% Similarly, JKNet~\cite{xu2018representation} designs a jumping knowledge network which fuses the features from different network layers. 
DropEdge~\cite{rong2019dropedge} proposes to randomly drop edges with a certain ratio before graph convolutions. These GNN models mainly focus on how to better enforce the \emph{smoothness} property. 
% In contrast, the proposed model additionally advocates to preserve the \emph{stability} and \emph{identifiability} against perturbations.
\textbf{Perturbations in GNN}: Employing perturbations to learn a robust GNN model has widely appeared in graph attack and defense. For example, both~\cite{xu2019topology} and~\cite{dai2018adversarial} explore the perturbations on edges to make attack and defense on graphs. 
In particular,~\cite{xu2019topology} analyzes the attack from an optimization perspective and proposes a gradient based model that can conduct both edge addition and deletion.~\cite{dai2018adversarial} explores the attacks by modifying the combinatorial
structure of data and further proposes a reinforcement learning based model that learns a generalizable attack policy.
% In order to further study the attack and defense on attributed graphs,~\cite{zugner2018adversarial} generates the adversarial perturbations targeting both the node attributes and structures. 
% More works study specific types of attack and defense such as the poisoning attack~\cite{tang2020transferring} and black-box attack~\cite{chang2020restricted}. 

Although SIGNNAP also employs perturbations, it is fundamentally different from the graph attack and defense works. 1). Graph attack aims to attack a target node and change its prediction labels, and graph defense targets to keep the prediction labels unchanged. 
While SIGNNAP targets to learn reliable node representations in an unsupervised manner. 
2). Graph attack involves techniques such as adversarial learning. While SIGNNAP is based on the theory of marginalizing noise and contrastive learning (CL).

\textbf{Contrastive Learning and GNN}: There are some parallel works incorporating contrastive learning with GNN. For instance, DGI~\cite{velickovic2018deep} maximizes the mutual information between a node representation and the high-level graph representation by contrastive learning. CMV~\cite{Kaveh2020Contrastive} extends DGI and builds contrastive signals between node representations and the graph representation in a multi-view formulation.
GMI~\cite{zhen2020graph} proposes to estimates the mutual information (MI) in DGI by decomposed parts. By incorporating random walking sampling and contrastive learning, GCC~\cite{qiu2020gcc} builds contrastive signals between sampled multi-view graphs. GRACE~\cite{Zhu_2020vf} corrupts both the graph structures and node attributes to generate two views and uses all other nodes from two augmentation views as negative pairs in the InfoNCE objective function. GCA~\cite{10.1145/3442381.3449802} designs adaptive augmentation on the graph topology and node attributes to incorporate various priors for topological and semantic aspects of the graph. Some works extends the negative-sampling free CL models from vision domain to graph domain. For example, BGRL~\cite{thakoor2021bootstrapped} extends BYOL~\cite{grill2020bootstrap}, a popular self-supervised learning method in computer vision, to graphs. 
BGRL includes carefully-designed tricks, such as stop gradient, non-symmetric networks and momentum encoders, to avoid degenerate solutions. 
Following Barlow Twins~\cite{zbontar2021barlow}, G-BT~\cite{bielak2021graph} utilizes a cross-correlation-based loss instead of the non-symmetric network in BGRL. There are some other works that concentrate on learning graph-level representations such as~\cite{LUO2023109448,DING2022108525}, which are different to our node-level learning problem and we would not introduce more here.

Although these methods also utilize contrastive learning, there are key differences between them and our SIGNNAP. DGI, CMV and GMI formalize the contrastive learning between a node representation and the graph representation. In this case, they try to keep more high-order information of nodes and achieve better \emph{smoothness}. Instead, SIGNNAP contrasts the representations of nodes, aiming to maintain the \emph{stability} and \emph{identifiability} against perturbations. \textcolor{black}{GCA specifically investigates adaptive augmentation strategies to construct positive and negative pairs in graph CL models, which is orthogonal to our work. The negative-sampling free graph CL models (i.e. BGRL and G-BT) usually suffer from degenerate solutions and thus require carefully designed learning tricks. By contrast, the learning process of CL with negatives samples is usually more stable and reliable to reach satisfying performance.
Compared to GCC, SIGNNAP has different motivations, methodology and different results, which would be demonstrated later.
% The differences are summarized in terms of the motivation, the methodology and the results. \textbf{1)}. In terms of the different motivations, GCC is motivated by pre-training of contrastive learning. SIGNNAP targets to learn node representations against perturbations by preserving the \emph{stability} and \emph{identifiability} properties.
% \textbf{2)}. In terms of different methodology, SIGNNAP enforces an unbiased \emph{stability} and \emph{identifiability} estimation by marginalizing the perturbation distribution on original samples, while the random walk sampling in GCC without analytic formulation may have large bias in estimation~\cite{ribeiro2010estimating}. \textbf{3)}. In terms of different results, SIGNNAP has better performance than GCC in the experiments. The effects of different perturbations on the input are also investigated.
}
\section{Method}
\begin{figure*}[t]
\centering
\includegraphics[width=12.0cm]{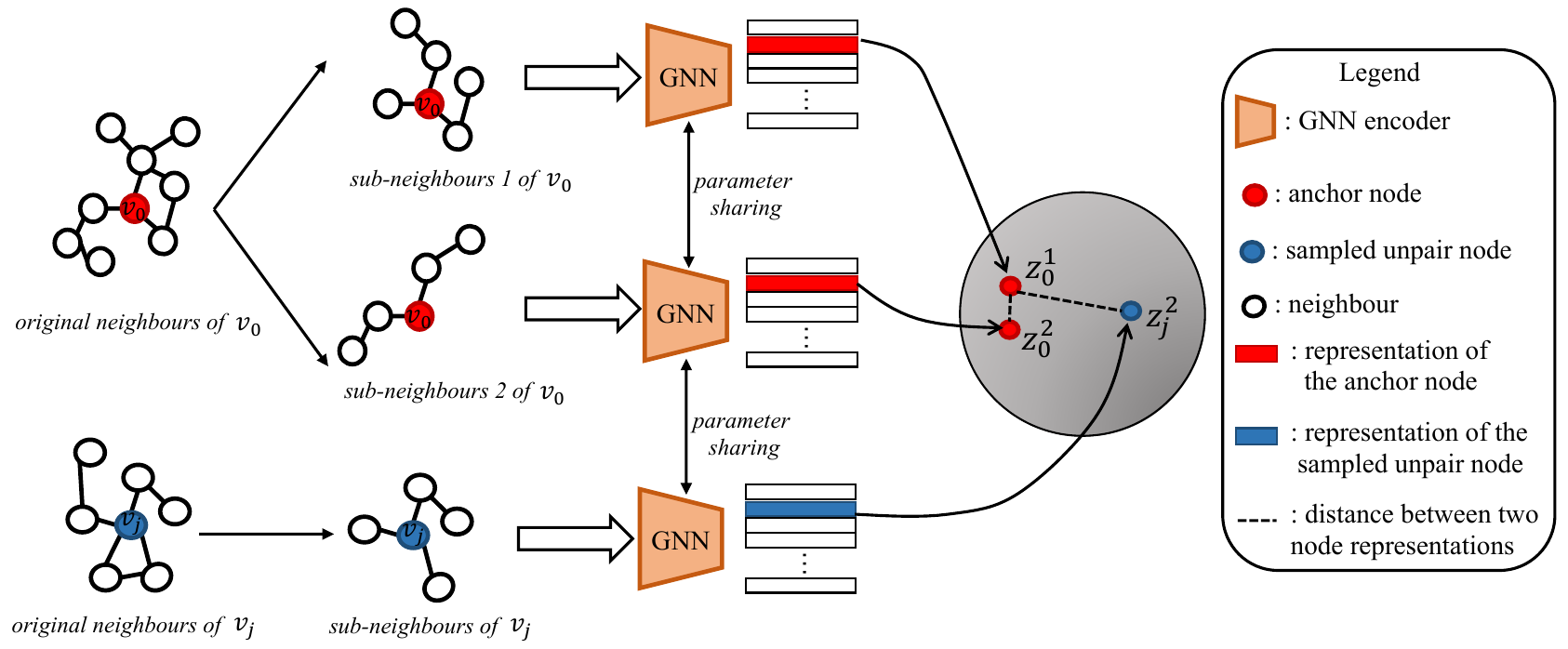}
% \vspace{-5pt}
\caption{The framework of our Stability-Identifiability GNN Against Perturbations (SIGNNAP).}
\label{figure:model_architecture}
\vspace{-10pt}
\end{figure*}

% \subsection{Problem Definition and Notation Description}
The problem of node representation learning on graphs is formulated as follows. Given an undirected graph $\mathcal{G}=(\mathcal{V}, A)$, $\mathcal{V}=\{v_{i}|1\leq i\leq N\}$ represents nodes in the graph and $A\in \mathbb{R}^{N\times N}$ denotes the adjacent matrix where $A_{ij}=1$ indicates node $i$ and node $j$ are connected while $A_{ij}=0$ means not.
When the nodes have attributes, we denote $X\in \mathbb{R}^{N\times F}$ as the attribute matrix, where $F$ is the attribute dimension. The goal of node representation learning on graphs is to encode nodes into representative embeddings. 
% The main notations used in this paper are summarized in Table~\ref{table:notations}. 
% \subsection{Overview}
% The proposed method learns reliable node representations against perturbations in terms of the \emph{stability} and \emph{identifiability}.
In SIGNNAP, we use popular GNN backbones to guarantee the nice \emph{smoothness} property. The \emph{stability} and \emph{identifiability} are maintained by a contrastive objective function. The general architecture of SIGNNAP is shown in Figure~\ref{figure:model_architecture}.

\subsection{Stability by Edge Perturbations}\label{sec:stability_by_edge}
To achieve the \emph{stability}, an intuitive way is to enforce a node representation from the slightly changed input to be close with the true one. However, it is non-trivial to obtain the true one. Based on~\cite{chen2015marginalizing}, we can achieve the \emph{stability} by marginalizing a perturbation distribution on the original samples. 

To clarify our method, given an arbitrary anchor node $v_{0}$ and a perturbation distribution $p(\epsilon)$, we can obtain the node representation $z_{0}$ by reparameterizing its true posterior $q(z_{0}|v_{0})$ with a differentiable transformation $f_{\theta}(\epsilon,v_{0})$ of a perturbation variable $\epsilon$:
\begin{align}
\label{eq:latent_code}
z_{0}=f_{\theta}(\epsilon, v_{0})~with~\epsilon\sim p(\epsilon)
\end{align}
In Eq.~\ref{eq:latent_code}, enforcing the \emph{stability} among multiple $z_{0}$ with different $\epsilon$ is computational-inefficient in each step. Instead, we employ an equivalent manner which enforces the \emph{stability} between every two random samples $z_{0}^{1},z_{0}^{2}$ from $f_{\theta}(\epsilon, v_{0})$ by sampling two $\epsilon$ independently. Since learning on graphs cares more about structures, here we use Bernoulli distribution on edges as $p(\epsilon)$. In particular, if we denote the original neighbours of $v_{0}$ as $\mathcal{N}_{0}$, we randomly drop edges of $\mathcal{N}_{0}$ with ratio $\rho$ to have different sub-neighbours of $v_{0}$. 
In this case, $p(\epsilon)$ is represented as $p(\epsilon;\rho)$.
Assuming $f_{\theta}$ is a GNN backbone that ensures the \emph{smoothness} property, we have:
\begin{equation}
\label{eq:GNN_encoder}
    z_{0}^{1}= f_{\theta}(\epsilon_{1},\mathcal{N}_{0}),z_{0}^{2}= f_{\theta}(\epsilon_{2},\mathcal{N}_{0})~with~\epsilon_{1},\epsilon_{2} \sim p(\epsilon;\rho)
\end{equation}
where $z_{0}^{1}$, $z_{0}^{2}$ are the anchor node's representations from two different sub-neighbours by corrupting $\mathcal{N}_{0}$ independently. Other perturbation distributions may have different effects. The effects of other common perturbations including perturbations on node attributes $X$ are studied in the experiments. Note that we mainly study perturbations with explicit distributions here to explore the general case of learning node embeddings against perturbations. 
More complicated types such as task-specific perturbations or generated perturbations are beyond the scope of this work and could be explored in future.

\subsection{Combating Perturbations by Stability and Identifiability}
In order to combat the perturbations for learning node representations, a typical way is to make sure a node representation remains stable under slight perturbations, i.e. the \emph{stability}. Besides, we propose to emphasize the original graph signals against perturbations by enforcing nodes with different structures are identifiable in latent space, i.e. the \emph{identifiability}. In our model, the \emph{stability} and \emph{identifiability} are interpreted as high similarity for representations of the same node with different perturbations and low similarity for representations of different nodes, respectively. Since contrastive learning has been widely proved as a good way to achieve the pair-similarity optimization, we propose to use a contrastive loss to formalize the two properties.
Specifically, we denote $(z_{0}^{1},
z_{0}^{2})$ as a positive pair. If
$z_{j}^{2}$ is a node representation not belonging to $v_{0}$ (i.e. $j\neq 0$), then $(z_{0}^{1}, z_{j}^{2})$ is a negative one.
Then the contrastive objective is shown as:
\begin{equation}
\label{eq:contrastive_loss1}
    \mathcal{L}_{1} = -\mathbb{E}_{\{z_{0}^{1},z_{0}^{2},z_{j}^{2}|_{j=1}^{K}\}} \Big [\log \frac{h_{\phi}(z_{0}^{1}, z_{0}^{2})}{\sum_{j=0}^{K}h_{\phi}(z_{0}^{1}, z_{j}^{2})} \Big ]
\end{equation}
where $h_{\phi}$ is a score function that is high for positive pairs and low for negative pairs. $K$ indicates the sampling size of negative pairs. Note that the expectation on $\{z_{0}^{1},z_{0}^{2},z_{j}^{2}|_{j=1}^{K}\}$ in Eq.~\ref{eq:contrastive_loss1} marginalizes the perturbation distribution in Eq.~\ref{eq:GNN_encoder}, which facilitates the unbiased \emph{stability} estimation.
Through the contrastive objective in Eq.~\ref{eq:contrastive_loss1}, we push the positive pairs $z_{0}^{1}$ and $z_{0}^{2}$ are closer than the negative pairs $z_{0}^{1}$ and $z_{j}^{2}$ in the embedding space. 
We implement the score function as the Gaussian potential kernel (also known as the Radial Basis Function (RBF) kernel).
\begin{equation}
\label{eq:critic_function}
    h_{\phi}(z_{0}^{1}, z_{0}^{2})=e^{{z_{0}^{1}}^{T}z_{0}^{2}/\tau}
\end{equation}
where $\tau$ is a temperature to control the distribution of $h_{\phi}$. Among a general class of kernels, the RBF kernel can well distribute the negative node representations uniformly in the embedding space for good \emph{identifiability}~\cite{cohn2007universally}. Besides, with the perturbations defined in Section~\ref{sec:stability_by_edge}, the \emph{stability} in SIGNNAP is equivalent to the isomorphism between two sub-graphs~\cite{hrushovski1992extending}. Through the above analysis, we conclude that the contrastive objective in our SIGNNAP matches the two properties well.
Similarly, by substituting the anchor representation $z_{0}^{1}$ in Eq.~\ref{eq:contrastive_loss1} with $z_{0}^{2}$, we have:  
\begin{equation}
\label{eq:contrastive_loss2}
    \mathcal{L}_{2} =-\mathbb{E}_{\{z_{0}^{2},z_{0}^{1},z_{j}^{1}|_{j=1}^{K}\}} \Big [\log \frac{h_{\phi}(z_{0}^{2}, z_{0}^{1})}{\sum_{j=0}^{K}h_{\phi}(z_{0}^{2}, z_{j}^{1})} \Big ]
\end{equation}
where $\mathcal{L}_{2}$ is a symmetric form of $\mathcal{L}_{1}$ and can help to stabilize the training.

\subsection{Objective Function}
By summing Eq.~\ref{eq:contrastive_loss1} and Eq.~\ref{eq:contrastive_loss2} up, we have our final objective function:
\begin{equation}
\label{eq:objective_function}
    \min_{\theta} \mathcal{L}=\mathcal{L}_{1}+\mathcal{L}_{2}
\end{equation}
where $\theta$ is the network parameter from the GNN backbone. Since the expectations in Eq.~\ref{eq:contrastive_loss1} and Eq.~\ref{eq:contrastive_loss2} do not have analytic formulations, we instead resort to unbiased Monte Carlo estimation to approximate the two equations. In the objective, $\rho$ and $K$ control the variations of \emph{stability} and \emph{identifiability} of the inputs, which is empirically analyzed with experiments in the experiments.
Moreover, we demonstrate that the proposed objective has connections with the mutual information between $z_{0}^{1}$ and $z_{0}^{2}$, which is shown as:
\begin{Lemma}
\label{lemma:lower_bound}
The proposed objective $\mathcal{L}$ is an estimator of the mutual information between $z_{0}^{1}$ and $z_{0}^{2}$, showing that:
\begin{equation}
\mathcal{I}(z_{0}^{1},z_{0}^{2})\geq \log(K)-\mathcal{L}   
\end{equation}
therefore minimizing $\mathcal{L}$ actually maximizes the \emph{lower bound} of $\mathcal{I}(z_{0}^{1},z_{0}^{2})$.
The \emph{lower bound} becomes tighter when $K$ becomes larger.
\end{Lemma}
Details about Lemma~\ref{lemma:lower_bound} is demonstrated in~\ref{appendix:lower_bound}.
Minimizing $\mathcal{L}$ maximizes the mutual information between $z_{0}^{1}$ and $z_{0}^{2}$ towards a direction where the positive pairs are more similar or dependent than the negative ones.

\subsection{Acceleration Strategies for Training}
\label{sec:acceleration_strategies}

\noindent\textbf{DropEdge Sampling Strategy}: 
If we denote the number of training epochs as $T$, the complexity of constructing various sub-neighbours of $N$ nodes is $\mathcal{O}(NT)$. In our implementation, following DropEdge,
we randomly drop edges on the adjacent matrix $A$ one time for each epoch and then feed the corrupted adjacent matrices $A^{1}$ and $A^{2}$ instead of $\mathcal{N}_{0}^{1}$ and $\mathcal{N}_{0}^{2}$ into the model training. Then the complexity is reduced to $\mathcal{O}(T)$ and dropping edge is accelerated. 
\textcolor{black}{To further accelerate training, like other contrastive loss based algorithms (e.g. DGI and CMV), we set the sampling method of negative pairs as the random sampling.}
\noindent\textbf{Memory Bank Strategy}:
In order to efficiently sample $K$ negative nodes for $\mathcal{L}$, we follow the memory bank strategy in~\cite{tian2019contrastive}. In particular, the latent features of all nodes are stored in memory and synchronously updated after loss back propagation. \textcolor{black}{It is also worthwhile to mention that the memory bank strategy here requires to save the representations of all nodes in learning, which may bring a memory challenge when the graph is extremely large. A dynamic memory queue for negative sampling may be a solution~\cite{he2020momentum}, and we will investigate how to combine it with SIGNNAP in future.}

\begin{table*}
\centering
\caption{The statistics of six benchmarks.}
% \vspace{-5pt}
\label{table:dataset_statistics}
\renewcommand{\arraystretch}{1.2}
\setlength{\tabcolsep}{0.9mm}{ 
\scalebox{0.9}{
\begin{tabular}{ccccccccc}
\hline
             & \#nodes & \#edges & \#density & \#classes & \#features & \#label rate & Train/Val/Test    \\ \hline
Pubmed       & 19,717  & 44,324  & 0.01\%    & 3         & 500        & 0.30\%        & 60/500/1,000      \\ \hline
Facebook     & 22,470  & 170,823 & 0.03\%    & 4         & 4,714      & 0.35\%        & 80/120/rest       \\ \hline
Coauthor-CS  & 18,333  & 81,894  & 0.02\%    & 15        & 6,805      & 1.60\%         & 300/450/rest      \\ \hline
Amazon-Com  & 13,752  & 245,861  & 0.13\%    & 10        & 767      & 1.45\%         & 200/300/rest      \\ \hline
Amazon-Pho  & 7,650  & 119,081  & 0.20\%    & 8        & 745      & 2.09\%         & 160/240/rest      \\ \hline
Coauthor-Phy & 34,493  & 247,962 & 0.02\%    & 5         & 8,415      & 57.98\%       & 20,000/5,000/rest \\ \hline
\end{tabular}
}}
\vspace{-10pt}
\end{table*}

\section{Experiments and Analysis}
\label{sec:experiments}
% In this section, we first demonstrate the experiment setups including dataset description, baselines and parameter settings. Then we show the performance comparison from different aspects in Section~\ref{sec:perform_comparison} and Section~\ref{sec:three_properties}. Finally, we explore the ablation studies in Section~\ref{sec:ablation}.
\subsection{Experiment Setups}
\label{sec:experiment_setup}
\subsubsection{Dataset Description}
We conduct the experiments on six benchmarks varying in graph types and sizes.\footnote{Note that Reddit dataset is quite large and is not included here since we do not have much computation resources. Instead, we include other benchmarks.} Pubmed is a widely used citation network. Facebook~\cite{rozemberczki2019multiscale} is a web page dataset where nodes are official Facebook pages and the edges are mutual connections between sites. Coauthor-CS and Coauthor-Phy~\cite{shchur2018pitfalls} are two coauthor datasets based on the Microsoft Academic Graph from the KDD Cup 2016 challenge. Amazon-Com (i.e. Amazon-Computer) and Amazon-Pho (i.e. Amazon-Photo) are two segments of the Amazon co-purchase graph~\cite{mcauley2015image}, where nodes are items and edges indicate two items are frequently co-purchased together.
Each dataset has raw node features and class labels, following mainstream works~\cite{perozzi2014deepwalk,Kaveh2020Contrastive}, we perform the node classification task to make evaluation. 
For Pubmed, Facebook, Coauthor-CS, Amazon-Com and Amazon-Pho, we conduct the transductive learning where all nodes and their raw features are accessible during training. For Coauthor-Phy, we conduct the inductive learning where the test nodes are not seen during training.

Different dataset splits on the node classification can have different evaluation values. The way of data splitting used in many graph CL works (i.e., DGI~\cite{velickovic2018deep},
CMV~\cite{Kaveh2020Contrastive}, GCC~\cite{qiu2020gcc}) originates from the semi-supervised works of graph representation learning.
% (i.e, GCN~\cite{kipf2016semi}, GAT~\cite{velickovic2017graph} and GraphSage~\cite{hamilton2017inductive}). 
Whereas some graph CL works, e.g., (GRACE~\cite{Zhu_2020vf}, BGRL~\cite{thakoor2021bootstrapped},
G-BT~\cite{bielak2021graph}, GCA~\cite{10.1145/3442381.3449802}), utilize a random splitting of the nodes into (80\%-10\%-10\%) train/validation/test set.
Usually, compared to that in the former
way, the latter way of splitting would favor the training, except for Coauthor-Phy in inductive learning setting. 
\textcolor{black}{In order to make comparison of different models, by following common graph learning works~\cite{velickovic2018deep,Kaveh2020Contrastive,qiu2020gcc}, we use the widely-recognized semi-supervised setting of data splitting. On Pubmed, the train/val/test nodes are the same as previous works~\cite{kipf2017semi,velickovic2018deep}. For Facebook, Coauthor-CS, Amazon-Com and Amazon-Pho, we follow the setting in~\cite{shchur2018pitfalls} where 20 nodes of each class are randomly sampled as the train set and 30 nodes of each class are randomly sampled as the validation set and the rest is the test set. 
For Coauthor-Phy in inductive learning, we randomly sample 20,000 nodes as train set and 5,000 nodes as validation set and the rest as test set.} 
The dataset statistics are shown in Table~\ref{table:dataset_statistics}.

\subsubsection{Baselines}
We make the comparison of our model with the following supervised models: GCN~\cite{kipf2017semi}, ResGCN~\cite{li2019deepgcns}, JKNet~\cite{xu2018representation}, GraphSage~\cite{hamilton2017inductive}, GAT~\cite{wang2019kgat},  DropEdge~\cite{rong2019dropedge} \textcolor{black}{and other unsupervised models: DeepWalk~\cite{perozzi2014deepwalk}, Node2vec~\cite{grover2016node2vec}, ARWMF~\cite{chen2019},
DGI~\cite{velickovic2018deep}, CMV~\cite{Kaveh2020Contrastive}, GCC~\cite{qiu2020gcc}, GRACE~\cite{Zhu_2020vf}, BGRL~\cite{thakoor2021bootstrapped}, G-BT~\cite{bielak2021graph} and GCA~\cite{10.1145/3442381.3449802}. 
% \textcolor{black}{According to the papers, GRACE, G-BT and GCA all use GCN as the encoder.}
Among the unsupervised models, BGRL and G-BT are learned without negatives while the others are not.}
% GCN, ResGCN, JKNet, GraphSage and GAT are also backbones of the proposed method.

\subsubsection{Parameter Settings}
We implement SIGNNAP with Pytorch on a machine with one Nvidia 1080-Ti GPU. For GraphSage, GAT and DGI, we use the codes from a famous GNN library- DGL~\cite{wang2019deep}.
For other baselines, we use the codes released by the authors.
% For ResGCN and JKNet with skip connections, we found 3 GNN-layer works best. For other GNN methods, the number of GNN layer is set as 2. 
% For GAT, we set 8 attention heads with each as 8 units. 
% The latent dimension is 128 for GCN, ResGCN, JKNet, GraphSage, DropEdge, DeepWalk and GCC, 200 for Node2Vec and ARWMF, 512 for DGI and CMV.
% ~\cite{rong2019dropedge,perozzi2014deepwalk}. 
% For Node2Vec and ARWMF, the latent dimension is 200.
% ~\cite{grover2016node2vec,chen2019}. 
% For DGI and CMV, the latent dimension is 512~\cite{velickovic2018deep,Kaveh2020Contrastive}. For GRACE, G-BT,BGRL and GCA, we follow the codes released by the authors. 
% The other hyper-parameters are set according to the original papers or codes. 
% For SIGNNAP, when incorporating different GNN methods as backbones, the number of GNN layer or latent dimension is the same as the backbones. 
% In other words, the number of GNN layer is 3 for SIGNNAP with ResGCN and JKNet as backbones, and 2 for other backbones. 
% For SIGNNAP with GAT as backbone, we have 8 attention heads with each as 8 units. 
% For other backbones, the latent dimension is set as 128. 
\textcolor{black}{In SIGNNAP, we construct the positive nodes with the same drop ratio $\rho=0.3$ like DropEdge~\cite{rong2019dropedge} and randomly sample $K=1024$ negative pairs.} We set the learning rate as 0.001 with 5,000 iterations and the temperature $\tau$ as 0.1.
% The drop ratio $\rho$ is set as 0.3, which is the same as DropEdge.
% By following~\cite{velickovic2018deep,qiu2020gcc,Kaveh2020Contrastive}, 
For unsupervised methods, we train a one-layer linear classifier for evaluation. 
% For all methods, the best trained classifier is chosen for testing according to the performance on the validation set. 
% Trainable parameters of all models are initialized with uniform distribution. 
More detailed parameter settings are summarized in~\ref{sec:params_setting}.

\begin{table*}[]
\caption{Classification accuracy($\%$) on different benchmarks. Standard deviation is reported in percentage format. 
The best value for unsupervised models is emphasized in bold. The second value for unsupervised models is emphasized with underline.
``-'' means the method does not support this setting. 
We also show the average rank of unsupervised models on the six benchmarks. \textcolor{black}{Lower rank score means better performance.} 
% In general, we recommend GraphSage as the backbone.
}
% \vspace{-5pt}
\label{table:overall_performance}
\centering
\renewcommand{\arraystretch}{1.0}
\setlength{\tabcolsep}{1.2mm}{ 
\scalebox{0.82}{
\begin{scriptsize} 
\begin{tabular}{c|ccccccc|c}
\hline
                               &                                           & \multicolumn{5}{c}{Transductive}                                                                                                     & Inductive                                  & \multirow{2}{*}{\begin{tabular}[c]{@{}c@{}}Average\\ rank\end{tabular}} \\ \cline{1-8}
                               & Method                                    & Pubmed                                     & Facebook                                   & Coauthor-CS                                &
                               Amazon-Com                                &
                               Amazon-Pho                                &
                               Coauthor-Phy                               &                                                                         \\ \hline
\multirow{10}{*}{Supervised}   & GCN                                       & 79.20($\pm$0.38)                           & 66.37($\pm$0.24)                           & 92.01($\pm$0.14)  &  81.18($\pm$0.27)        &  85.82($\pm$0.30)             & 93.35($\pm$0.02)                           & \multirow{10}{*}{/}                                                     \\ \cline{2-8}
                               & ResGCN                                    & 77.74($\pm$0.39)                           & 67.69($\pm$0.60)                           & 92.84($\pm$0.24)                           &
                               81.10($\pm$0.70)                           &
                               87.21($\pm$0.50)                           &
                               95.88($\pm$0.03)                           &                                                                         \\ \cline{2-8}
                               & JKNet                                     & 77.84($\pm$0.11)                           & 68.09($\pm$0.75)                           & 92.76($\pm$0.22)                           &
                               80.91($\pm$0.83)                           &
                               87.25($\pm$0.50)                           &
                               95.56($\pm$0.15)                           &                                                                         \\ \cline{2-8}
                               & GraphSage                                 & 79.02($\pm$0.31)                           & 69.62($\pm$0.38)                           & 92.60($\pm$0.16)                           & 
                               82.10($\pm$0.22)                           &
                               87.60($\pm$0.34)                           &
                               95.28($\pm$0.16)                           &                                                                         \\ \cline{2-8}
                               & GAT                                       & 78.71($\pm$0.21)                           & 72.24($\pm$0.07)                       & 91.23($\pm$0.06)                           & 
                               81.85($\pm$0.73)                           &
                               87.02($\pm$1.51)                           &
                               93.96($\pm$0.04)                           &                                                                         \\ \cline{2-8}
                               & DropEdge(GCN)                             & 78.82($\pm$0.29)                           & 66.10($\pm$0.19)                           & 92.12($\pm$0.12)                           & 
                               81.42($\pm$0.21)                           &
                               85.58($\pm$0.13)                           &
                               93.32($\pm$0.02)                           &                                                                         \\ \cline{2-8}
                               & DropEdge(ResGCN)                          & 77.52($\pm$0.38)                           & 67.30($\pm$0.75)                           & 93.01($\pm$0.16)                       & 
                               81.50($\pm$0.74)                           &
                               87.46($\pm$0.27)                           &
                               95.89($\pm$0.04)                       &                                                                         \\ \cline{2-8}
                               & DropEdge(JKNet)                           & 77.85($\pm$0.16)                           & 67.68($\pm$0.54)                           & 92.74($\pm$0.20)                           & 
                               81.70($\pm$0.43)                           &
                               87.01($\pm$0.42)                           &
                               95.77($\pm$0.02)                           &                                                                         \\ \cline{2-8}
                               & DropEdge(GraphSage)                       & 78.53($\pm$0.31)                           & 69.25($\pm$0.30)                           & 92.77($\pm$0.08)                           & 
                               82.11($\pm$0.30)                           &
                               87.64($\pm$0.40)                           &
                               95.41($\pm$0.30)                           &                                                                         \\ \cline{2-8}
                               & DropEdge(GAT)                             & 78.90($\pm$0.30)                           & 71.57($\pm$0.21)                           & 91.32($\pm$0.14)                           & 
                               82.20($\pm$0.50)                           &
                               87.59($\pm$0.73)                           &
                               93.80($\pm$0.13)                           &                                                                         \\ \hline
\multirow{11}{*}{Unsupervised} & DeepWalk                                  & 65.59($\pm$1.43)                           & 63.04($\pm$1.32)                           & 77.81($\pm$0.86)                           & 
76.93($\pm$0.73)                           &
81.50($\pm$0.82)                           &
91.17($\pm$0.33)                           & 9.83                                                                    \\ \cline{2-9} 
                               & Node2Vec                                  & 70.34($\pm$0.69)                           & 69.69($\pm$0.77)                           & 79.93($\pm$0.62)                           & 
                               75.49($\pm$1.08)                           &
                               82.21($\pm$0.67)                           &
                               91.43($\pm$0.73)                           & 10.0                                                                    \\ \cline{2-9} 
                               & ARWMF                                     & 78.03($\pm$0.85)                           & 61.97($\pm$0.94)                           & 86.02($\pm$0.82)                           & 
                               68.34($\pm$1.54)                           &
                               78.35($\pm$0.56)                           &
                               -                                          & -                                                                    \\ \cline{2-9} 
                               & DGI                                       & 79.24($\pm$0.50)                           & 69.53($\pm$1.25)                           & 91.41($\pm$0.12)                           & 
                               71.41($\pm$0.85)                           &
                               79.34($\pm$0.66)                           &
                               93.26($\pm$0.35)                           & 8.66                                                                    \\ \cline{2-9} 
                               & CMV                                       & 80.10($\pm$0.10)                           & 67.24($\pm$0.13)                           & 90.73($\pm$0.71)                           & 
                               67.15($\pm$1.98)                           &
                               79.54($\pm$1.47)                           &
                               91.49($\pm$0.41)                           & 9.83                                                                     \\ \cline{2-9} 
                               & GCC                                       & 80.60($\pm$0.45)                       & 70.36($\pm$0.53)                       & 91.76($\pm$0.12)                       & 
                               74.18($\pm$1.02)                           &
                               83.60($\pm$0.52)                           &
                               \textbf{93.97($\pm$0.16)} & 5.5                                                                    
                               \\ \cline{2-9}
                              & GRACE                                       & 78.10($\pm$0.10)                       & 65.81($\pm$0.27)                       & 90.20($\pm$0.33)                       & 
                               69.32($\pm$0.78)                           &
                               66.48($\pm$0.30)                           &
                               72.90($\pm$0.18) & 12.16                                                                    
                               \\ \cline{2-9}
                             & BGRL                                       & 71.01($\pm$0.42)                       & 62.42($\pm$0.21)                       & 89.31($\pm$0.45)                       & 
                               \textbf{81.72($\pm$0.86)}                           &
                               86.02($\pm$0.53)                           &
                               75.85($\pm$0.20) & 9.0                                                                    
                               \\ \cline{2-9}
                             & G-BT                                       & 80.04($\pm$0.63)                       & 64.14($\pm$0.47)                       & 90.38($\pm$0.62)                       & 
                               75.17($\pm$1.05)                           &
                               84.49($\pm$0.66)                           &
                               74.05($\pm$0.25) & 8.83                                                                    
                               \\ \cline{2-9}
                             & GCA                                       & 80.02($\pm$0.18)                       & 71.52($\pm$0.34)                       & 90.97($\pm$0.29)                       & 
                               77.30($\pm$0.58)                           &
                               85.50($\pm$0.40)                           &
                               73.09($\pm$0.21) & 6.5                                                                    
                               \\ \cline{2-9}
                               
                               & \textbf{SIGNNAP(JKNet)}     & 80.82($\pm$0.48)                           & 68.77($\pm$1.31)                           & 91.48($\pm$0.18)  &       73.91($\pm$0.50)                           &
                               83.55($\pm$0.29)                           &
                                92.82($\pm$0.08)                           & 7.0
                                         
                                         \\ \cline{2-9}

                               & \textbf{SIGNNAP(GAT)}       & 80.93($\pm$0.40)                           & \textbf{78.39($\pm$0.64)} & 90.14($\pm$0.37) &   75.11($\pm$0.47)                           &
                               \textbf{89.84($\pm$0.26)}                           &
                                92.34($\pm$0.03)                           & 4.5                
                               \\ \cline{2-9} 
                               & \textbf{SIGNNAP(ResGCN)}    & 77.34($\pm$0.53)                           & 71.21($\pm$0.29)                           & \underline{92.01}($\pm$0.19)  &       77.24($\pm$0.28)                           &
                               84.09($\pm$0.21)                           &
                                \underline{93.43}($\pm$0.27)                       & 5.16                                                                                                                                           \\
                                \cline{2-9} 
                               & \textbf{SIGNNAP(GCN)}       & \underline{81.34}($\pm$0.54)                           & 71.13($\pm$0.16)                           & \textbf{92.35($\pm$0.27)}&
                               74.65($\pm$0.16)                           &
                               85.74($\pm$0.25)                           &
                                92.76($\pm$0.10)                           & 4.66                                                               \\ \cline{2-9} 
                               & \textbf{SIGNNAP(GraphSage)} & \textbf{81.81($\pm$0.20)} & \underline{71.55}($\pm$0.48)                           & 91.21($\pm$0.24) &     \underline{78.49}($\pm$0.08)                           &
                               \underline{86.26}($\pm$0.43)                           &
                                93.33($\pm$0.02)                           & \textbf{2.83}                                                                                                                 \\ \cline{1-9} 
\end{tabular}
\end{scriptsize}
}}
\vspace{-10pt}
\end{table*}
\subsection{Performance Comparison}
\label{sec:perform_comparison}
\subsubsection{Node Classification Performance}
\label{sec:node_classification}
Following mainstream works~\cite{perozzi2014deepwalk,tang2015line,qiu2018network,velickovic2018deep,Kaveh2020Contrastive}, we conduct the node classification task to verify the effectiveness of the proposed method.
The results on different benchmarks are summarized in Table~\ref{table:overall_performance}. In this table, we report the mean classification accuracy (with standard derivation) on the test nodes after 5 runs with different random seeds. From the table, we have the following observations.

\textbf{1)}.The proposed method generally performs better than recent unsupervised models and even exceeds the supervised models in some cases. 
On Pubmed, SIGNNAP(GCN) reaches a 2.10\% gain over DGI and a 0.74\% gain over GCC.
% On Facebook, SIGNNAP(GAT) has a 8.86\% gain over DGI and a 8.03\% gain over GCC.
\textcolor{black}{With the same encoder, SIGNNAP(GCN) can generally achieve a better rank score of 4.66, compared to GRACE with 12.16, G-BT with 8.83, GCA with 6.5 and GCC with 5.5,
which illustrates that the improvement is from the model itself instead of the backbone.}
% \textcolor{black}{Compared to GRACE, G-BT and GCA, SIGNNAP(GCN) uses the same encoder and generally has a better rank score of 4.66, which can illustrate SIGNNAP's better representation learning ability.}
% On Amazon-Com, SIGNNAP(GraphSage) reaches a 4.31\% gain over GCC and a 7.08\% gain over DGI. 
% \textbf{2)}.On Coauthor-CS and Amazon-Pho, the proposed unsupervised method shows competitive performance compared to the supervised models and performs better than other unsupervised models.
\textbf{2)}.For the inductive learning on Coauthor-Phy, we can see both DGI, GCC and SIGNNAP perform a bit worse than the supervised methods. \textcolor{black}{A possible reason is that the label rate of Coauthor-Phy is quite large compared to other datasets. The supervised models are not easy to be over-fitting with a large label rate.}
\textcolor{black}{\textbf{3)}. It is interesting to find that SIGNNAP(GraphSage) not always performs better than SIGNNAP(GCN). This is 
understandable since different datasets have different distributions that favor different backbones. 
% This can also be verified through the results on GCN, ResGCN, JKNet, GraphSage and GAT in Table~\ref{table:overall_performance}. 
% For instance, GraphSage reaches the best performance on Amazon-Com and Amazon-Pho but not the best performance on other datasets. 
In our experiments, we use these common backbones to demonstrate SIGNNAP's flexibility in use.
% From the average rank score, we see that SIGNNAP(GraphSage) has a better performance and generally we recommend GraphSage as the backbone.
}
\begin{figure*}[ht]
%第一行 loss
\centering
\begin{minipage}[t]{\textwidth}
\centering
\includegraphics[width=\textwidth]{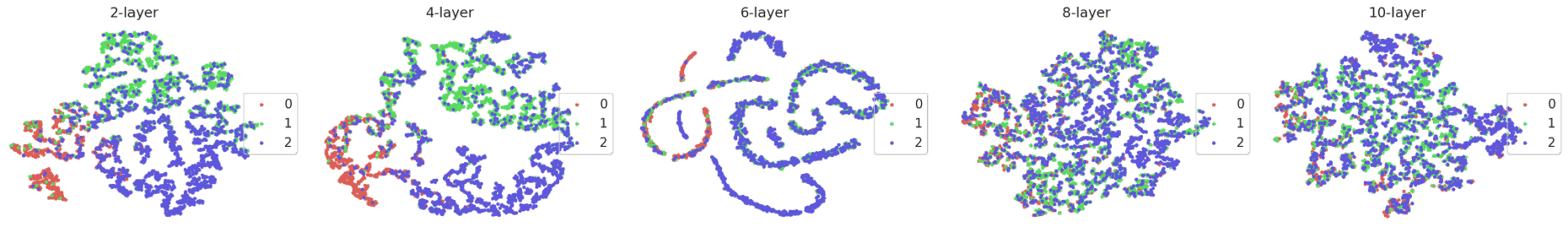}
% \vspace{-15pt}
\caption*{(a) GCN}
\end{minipage}\\
\begin{minipage}[t]{\textwidth}
\centering
\includegraphics[width=\textwidth]{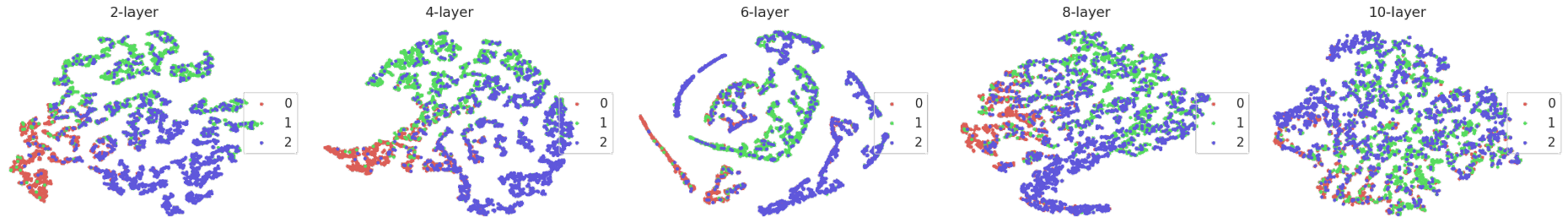}
% \vspace{-15pt}
\caption*{(b) DropEdge(GCN)}
\end{minipage}\\
\begin{minipage}[t]{\textwidth}
\centering
\includegraphics[width=\textwidth]{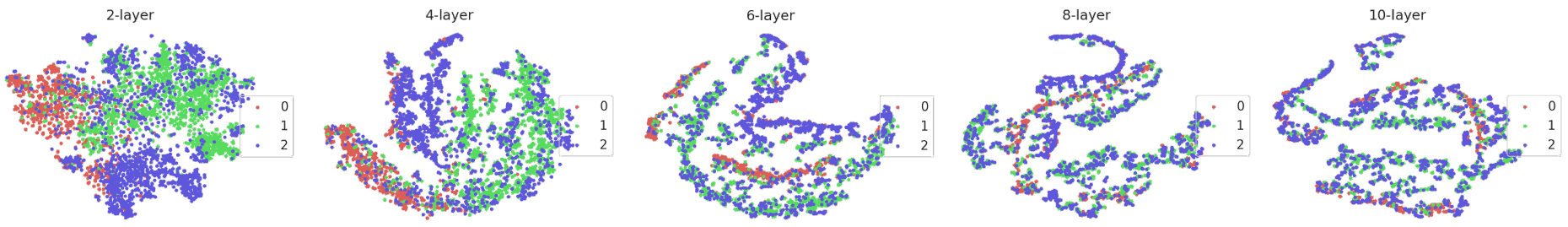}
% \vspace{-15pt}
\caption*{(c) DGI}
\end{minipage} \\
\begin{minipage}[t]{\textwidth}
\centering
\includegraphics[width=\textwidth]{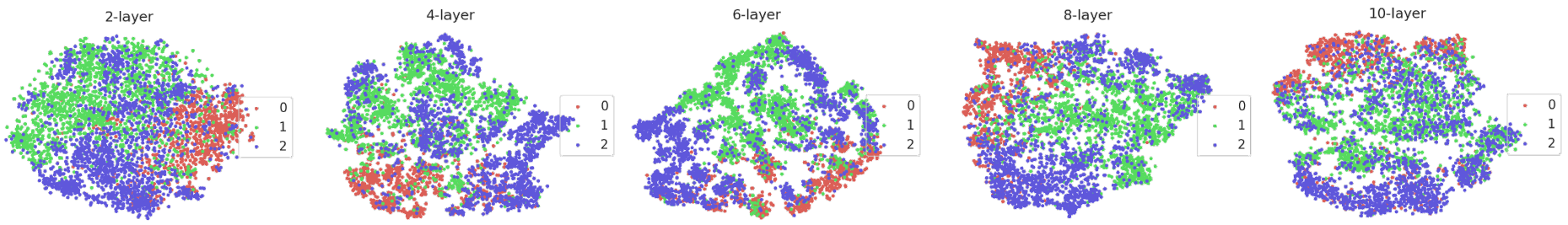}
% \vspace{-15pt}
\caption*{(d) SIGNNAP(GCN)}
\end{minipage} \\
% \vspace{-5pt}
\caption{The results of over-smoothing with different number of GCN layers on Pubmed. \textcolor{black}{The number of layers is 2,4,6,8,10 from left to right in each row.} We show the results of four models here to give an example. Different colors indicate different categories. Results of other methods and datasets follow similar patterns.}
\label{figure:over_smoothing}
\vspace{-10pt}
\end{figure*}
\subsubsection{On Preventing Over-smoothing}
As discussed in~\cite{li2018deeper}, when stacking too many GCN layers, the over-smoothing issue arises which means the top-layer node embeddings converge to the same subspace and become unidentifiable from each other. As SIGNNAP enforces the \emph{identifiability} property which tries to learn different embeddings for nodes with different structures, it is curious to see if SIGNNAP has better ability of preventing the over-smoothing issue. In this experiment, we visualize the node embeddings by t-SNE when increasing the depth of GCN layers. The results of different models are shown in Figure~\ref{figure:over_smoothing}. 

From this figure, we can see that: \textbf{1)}. When increasing the number of GCN layers (e.g. 10-layer), SIGNNAP(GCN) shows more identifiable node embeddings while baseline models easily have collapsed subspace. 
% \textbf{2)}. Compared to DGI that highlights the global features, SIGNNAP(GCN) resorts to maintain the \emph{identifiability} by a contrastive learning objective and can help to prevent the over-smoothing problem. 
\textbf{2)}. DropEdge(GCN) can alleviate the over-smoothing problem in GCN, while it does not perform better than SIGNNAP(GCN). The contrastive objective in SIGNNAP(GCN) can work with DropEdge together and show better performance. \textbf{3)}. Note that the embeddings of GCN and DropEdge(GCN) are learned in a supervised manner while those of DGI and SIGNNAP(GCN) are learned in an unsupervised manner, so the supervised models may have better performance than the unsupervised ones when the number of layers is 2 or 4.

\begin{figure*}[ht]
%第一行 loss
\centering
\begin{minipage}[t]{0.3\textwidth}
\centering
\includegraphics[width=\textwidth]{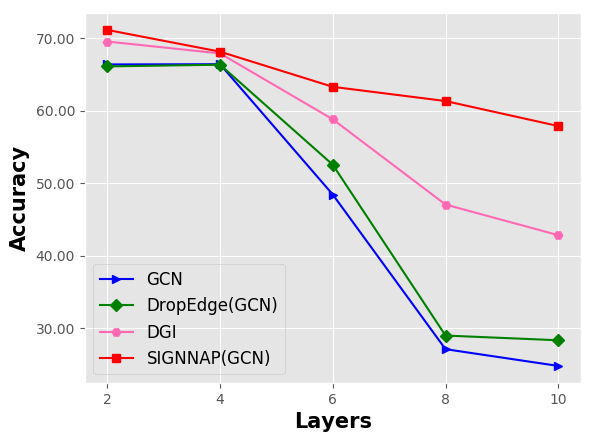}
% \vspace{-15pt}
\caption*{(a) Different training layers}
\end{minipage}
\begin{minipage}[t]{0.3\textwidth}
\centering
\includegraphics[width=\textwidth]{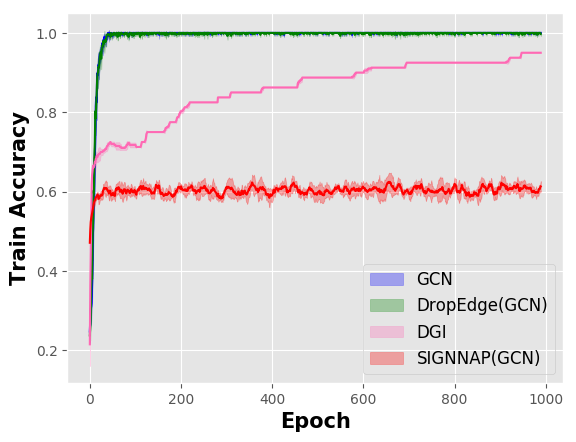}
% \vspace{-15pt}
\caption*{(b) Train accuracy}
\end{minipage}
\begin{minipage}[t]{0.3\textwidth}
\centering
\includegraphics[width=\textwidth]{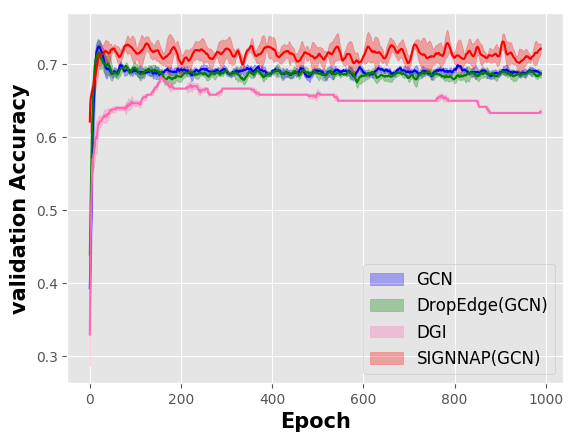}
% \vspace{-15pt}
\caption*{(c) Validation accuracy}
\end{minipage} \\
\vspace{-5pt}
\caption{The results of over-fitting on Facebook. (a) shows the comparison results of different training layers. (b) indicates the train accuracy of different methods. (c) indicates the validation accuracy of different methods. The accuracy curves of unsupervised models are plot based on the one-layer linear classifier. We show the results of several baselines here to illustrate the idea. Results of other methods and datasets follow similar patterns.}
\label{figure:training_curve}
\vspace{-10pt}
\end{figure*}
\subsubsection{On Preventing Over-fitting}
In this experiment, we investigate the over-fitting problem of different models. We show the results of different training layers, train accuracy and validation accuracy of different methods in Figure~\ref{figure:training_curve}.

By analyzing the results in Figure~\ref{figure:training_curve}, we make the following summarization: \textbf{1)}. From Figure~\ref{figure:training_curve} (a), \textcolor{black}{we see that the proposed SIGNNAP has improvement over baseline models. Especially when the number of layer increases from 2 to 10, with nearly four times increased learning parameters, the accuracy of baselines drops from 67\% to nearly 30\%, while the accuracy of SIGNNAP(GCN) drops from 72\% to nearly 60\%.} 
% This indicates that SIGNNAP(GCN) has better ability on preventing the over-fitting problem. 
\textbf{2)}. Besides, from Figure~\ref{figure:training_curve} (b) and (c), we observe that the supervised models (GCN and DropEdge) have a severe over-fitting problem. 
% The unsupervised models DGI and SIGNNAP do not easily overfit the data. 
For SIGNNAP, we can see that although it has the lowest train accuracy, it shows the best validation accuracy. 
It is mainly because SIGNNAP learns more stable node representations, namely the GNN feature extractor has less variance and has better generalization ability.
% \textbf{3)}. It is also worthwhile to point out that there are other works trying to solve the over-fitting problem on graphs~\cite{rong2019dropedge,li2019deepgcns,xu2018representation}. These methods are orthogonal to ours and we can incorporate them together. For example, we use ResGCN~\cite{li2019deepgcns} and JKNet~\cite{xu2018representation} as backbones and DropEdge~\cite{rong2019dropedge} as our edge dropping method.

\begin{figure*}[ht]
%第一行 loss
\centering
% \begin{minipage}[t]{0.22\textwidth}
% \centering
% \includegraphics[width=\textwidth]{images/GCN_facebook_page_contunuity_matrix.png}
% \vspace{-15pt}
% \caption*{(a) GCN}
% \end{minipage}
\begin{minipage}[t]{0.24\textwidth}
\centering
\includegraphics[width=\textwidth]{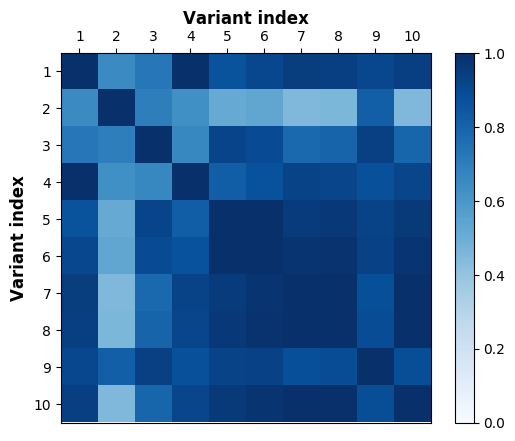}
% \vspace{-15pt}
\caption*{(a) DropEdge(GCN)}
\end{minipage}
\begin{minipage}[t]{0.24\textwidth}
\centering
\includegraphics[width=\textwidth]{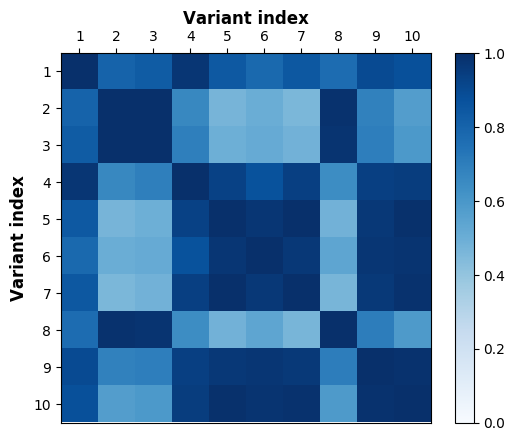}
% \vspace{-15pt}
\caption*{(b) DGI}
\end{minipage}
\begin{minipage}[t]{0.24\textwidth}
\centering
\includegraphics[width=\textwidth]{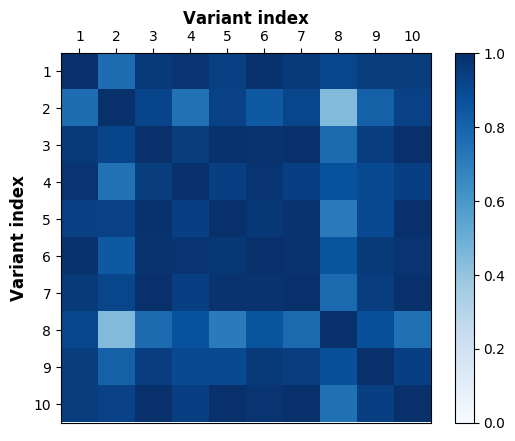}
% \vspace{-15pt}
\caption*{(c) GCC}
\end{minipage}
\begin{minipage}[t]{0.24\textwidth}
\centering
\includegraphics[width=\textwidth]{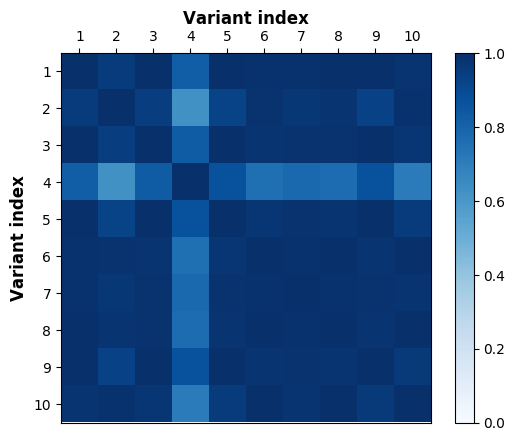}
% \vspace{-10pt}
\caption*{(d) SIGNNAP(GCN)}
\end{minipage}\\
% \vspace{-10pt}
\caption{The representation \emph{stability} comparison on Facebook. The mean value of the normalized cosine similarity in (a) (b) (c) (d) respectively is 0.857, 0.804, 0.915 and 0.946. Deeper color indicates larger similarity and more stability against perturbations on the input.} 
% Results of several baselines are shown here to provide clear Comparison and better illustrate our idea. Results of other methods follow similar patterns.}
\label{figure:contunuity_comparison}
\vspace{-10pt}
\end{figure*}
\subsection{Empirical Results of the Three Properties}\label{sec:three_properties}
\subsubsection{Stability of Node Representations}
Here we conduct an experiment to show that our method learns more stable node representations on graphs. Specifically, after we train a model, we fix the model and add perturbations to the input by dropping edges with $\rho=0.3$ ten times. In this way, for an arbitrary node $v_{0}$, we can have its ten variant inputs and their corresponding ten node representations by the fixed model.
% Then, we can output 10 node representations of the 10 variant inputs by the fixed model. 
Next, we calculate the cosine similarity matrix $S\in \mathbb{R}^{10\times 10}$ of these ten variants' representations. If a model learns stable node representations, the similarity between each two representations should be large. Here, we sample the test nodes and calculate the $S$ matrix of each node for four methods. For each method, we average the $S$ matrices of test nodes to obtain one matrix denoted as $\overline{S}$. For better comparison, we apply min-max normalization on the $\overline{S}$ matrices and show the results in Figure~\ref{figure:contunuity_comparison}.

From Figure~\ref{figure:contunuity_comparison}, we summarize that the proposed method learns more stable node representations. \textbf{1)}.DGI have many light-colored blocks, which means the node representation is sensitive to the slight changes on the input. 
\textbf{2)}.Compared to DGI, DropEdge(GCN) shows a better result with deeper-colored blocks, because DropEdge(GCN) tries to maintain the \emph{stability} by assigning the same label to the input with slight perturbations. When using the label as an agent, it is hard to guarantee the \emph{stability} on the node representations directly.
Instead, the proposed SIGNNAP explicitly imposes the \emph{stability} on learned node representations. 
\textbf{3)}.Compared to GCC, SIGNNAP shows better \emph{stability} results, because SIGNNAP can guarantee an unbiased \emph{stability} estimation by marginalizing the perturbation distribution. While the random walk sampling in GCC without analytic formulation may have large bias in learning \emph{stability}. 
% Besides, the random walk sampling requires a large number of walks for high-quality statistics, while DropEdge in SIGNNAP is more efficient~\cite{rong2019dropedge}.
% Thereby, SIGNNAP(GCN) shows the best results, and almost all blocks in Figure~\ref{figure:contunuity_comparison} (d) are deep-colored. 

\begin{figure*}[ht]
%第一行 loss
\centering
\begin{minipage}[t]{0.24\textwidth}
\centering
\includegraphics[width=\textwidth]{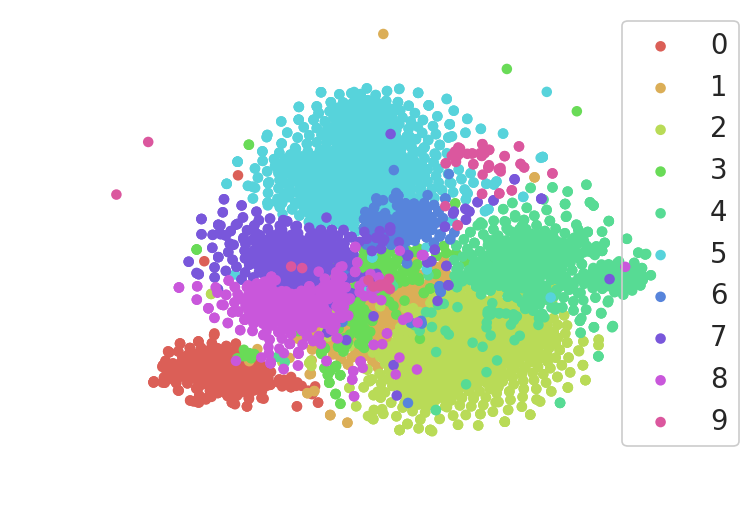}
% \vspace{-15pt}
\caption*{(a) Raw}
\end{minipage}
\begin{minipage}[t]{0.24\textwidth}
\centering
\includegraphics[width=\textwidth]{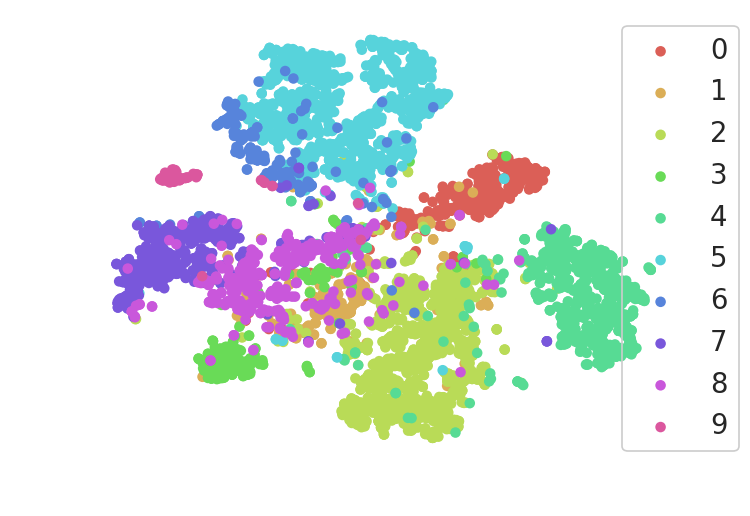}
% \vspace{-15pt}
\caption*{(b) DGI}
\end{minipage}
\begin{minipage}[t]{0.24\textwidth}
\centering
\includegraphics[width=\textwidth]{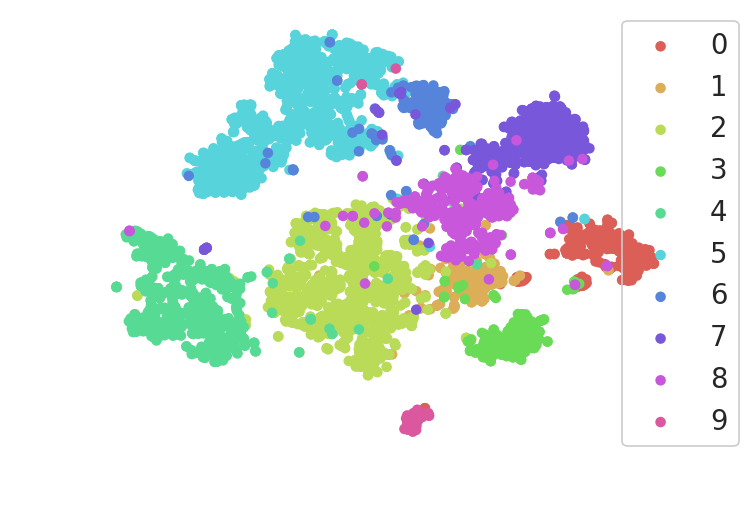}
% \vspace{-15pt}
\caption*{(c) GCC}
\end{minipage}
\begin{minipage}[t]{0.24\textwidth}
\centering
\includegraphics[width=\textwidth]{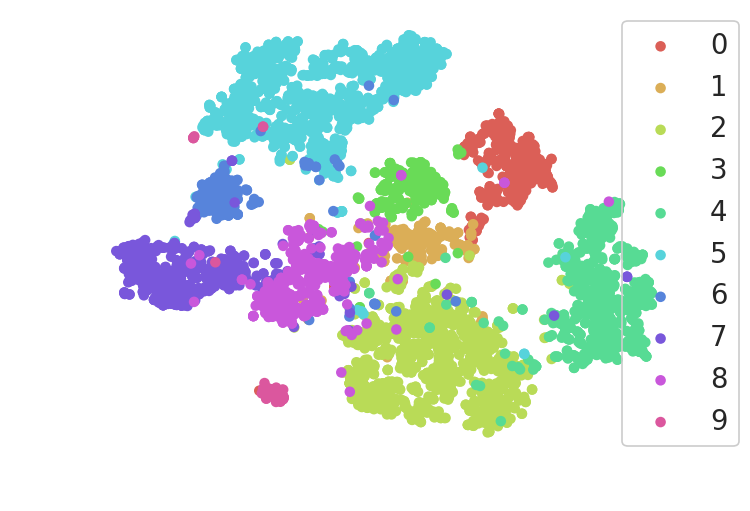}
% \vspace{-15pt}
\caption*{(d) SIGNNAP(GCN)}
\end{minipage} \\
% \vspace{-10pt}
\caption{The t-SNE comparison of learned node representations on Coauthor-CS. Note that for clear presentation, we visualize the first ten categories of nodes. (a) ``Raw'' means the raw node features are used. (b) (c) and (d) indicate the features are learned by DGI, GCC and SIGNNAP(GCN), respectively.
% The Silhouette score for (a) (b) (c) (d) respectively is -0.0071, 0.2200, 0.2234, 0.2269.
}
% We show results of three methods here to make a clear comparison and better illustrate our idea. Results of other methods follow similar pattern.}
\label{figure:tsne_comparison}
\vspace{-10pt}
\end{figure*}

\subsubsection{Smoothness and Identifiability of Node Representations}
\label{appen:tsne_comparison}
We also conduct an experiment to show the \emph{smoothness} and \emph{identifiability} of learned node representations. 
Discussing \emph{smoothness} and \emph{identifiability} without any data is hard to understand. 
Here we use node categories to illustrate this by assuming that nodes from the same category should be clustered together (i.e. \emph{smoothness}) and nodes from different categories should be discriminative (i.e. \emph{identifiability}). In particular, we visualize the learned node representations on Coauthor-CS by t-SNE in Figure~\ref{figure:tsne_comparison}. 

From Figure~\ref{figure:tsne_comparison} (a), it is clear that raw features are easily overlapped together. DGI, GCC and SIGNNAP(GCN) show better performance by identifying nodes from different clusters. 
\textcolor{black}{Although GCC has similar loss formulation as SIGNNAP, it still shows inferior performance to SIGNNAP(GCN). We consider it is caused by the differences between them. \textbf{1)}.
\textcolor{black}{GCC aims to learn the structural similarity and transferability in latent space by a pertaining task, so as to benefit downstream tasks. By contrast, SIGNNAP concentrates on the model's robust representation learning under different perturbations.}
% GCC is motivated by pre-training of contrastive learning. SIGNNAP targets to learn node representations against perturbations by preserving the \emph{stability} and \emph{identifiability} properties.
\textbf{2)}. SIGNNAP enforces an unbiased \emph{stability} and \emph{identifiability} estimation by marginalizing the perturbation distribution on original samples, while the random walk sampling in GCC without analytic formulation may have large bias in estimation~\cite{ribeiro2010estimating}. 
% \textbf{3)}. In terms of different results, SIGNNAP has better performance than GCC in the experiments. The effects of different perturbations on the input are also investigated.
The better performance of SIGNNAP over GCC can also be verified in Table~\ref{table:overall_performance}.}
% Generally, SIGNNAP(GCN) shows the most clustered and discriminative node representations. This indicates node representations of SIGNNAP are smooth in the same cluster and are identifiable among different clusters.

\subsection{Ablation Study}
\subsubsection{Different Perturbations}
\label{sec:ablation}
% \subsubsection{Different Perturbations}
In the method part, we use DropEdge which is an edge removal strategy to obtain different samples. We do not consider edge addition here since it has a complexity of $\mathcal{O}(N^{2})$ and will incur much more noise than edge removal. For any model, proper noise in training can help it; while overwhelming noise level leads to deterioration. How to add edges is an interesting topic~\cite{xu2019topology} but is beyond the scope of SIGNNAP here.

In addition, we can also add perturbations on node attributes to have different samples of an anchor node $v_{0}$. Thus, we further explore the effects of different perturbations on node attributes. In particular, if we denote the attribute vector of $v_{0}$ as $x_{0}$ and a perturbation vector as $\delta$, then the corrupted attribute vector can be represented as $x_{0}^{'}=x_{0}+\delta$,
where $\delta$ can be sampled from different distributions such as Gaussian distribution. We show the results of different perturbations in Table~\ref{table:diff_perturbations}. Note that we multiply 0.01 for noise from these distributions and then denote them as $\delta$ to adapt to the scale of normalized features. \textcolor{black}{Here, we fix the backbone as GCN for SIGNNAP to analyze the effects of different perturbations.}
\begin{table*}[]
\caption{The results of SIGNNAP(GCN) with different kinds of perturbations. ``A'', ``X'' and ``A+X'' respectively indicate the perturbations on structures, node attributes or both. ``Gaussian'', ``Laplace'' and ``Uniform'' respectively indicate perturbations that are sampled from $N(0,1)$, $L(0,1)$, $U(0,1)$.}
% \vspace{-5pt}
\label{table:diff_perturbations}
\centering
\renewcommand{\arraystretch}{1.2}
\setlength{\tabcolsep}{0.5mm}{ 
\scalebox{0.8}{
\begin{tabular}{c|c|cccccc}
\hline
Type                                                            & Approach                                                     & Pubmed & Facebook & Coauthor-CS
& Amazon-Com
& Amazon-Pho
& Coauthor-Phy \\ \hline
A                                                               & DropEdge                                                     & \textbf{81.34}  & 71.13    & \textbf{92.35} &\textbf{74.65}   &85.74   & 92.76        \\ \hline
\multirow{3}{*}{X}                                              & Gaussian                                                     & 77.82  & 70.70        & 90.11
&74.17   &\textbf{86.26}
& 92.03            \\ \cline{2-8} 
                                                                & Laplace                                                      & 76.98  & 71.22        & 89.25
                                                                &72.00   &85.43
                                                                & 92.24            \\ \cline{2-8} 
                                                                & Uniform                                                      & 77.48  & 67.04        & 87.88
                                                                &67.67   &83.20
                                                                & 92.11            \\ \hline
\multirow{3}{*}{A+X} & DropEdge+Gaussian & 80.02  & 71.15    & 91.06  &72.84   &85.64
& \textbf{92.91}        \\ \cline{2-8} 
                                                                & DropEdge+ Laplace  & 77.58  & \textbf{71.38}    & 90.56      
                                                                &72.05   &86.00
                                                                & 92.69        \\ \cline{2-8} 
                                                                & DropEdge+Uniform  & 77.68      & 69.67        & 89.48    
                                                            &58.65   &76.54    
                                                                & 92.60            \\ \hline
\end{tabular}
}}
\vspace{-10pt}
\end{table*}

From Table~\ref{table:diff_perturbations}, we can see: \textbf{1)}.\textcolor{black}{Comparing perturbations on ``A'' and ``X'', the result of `X`'' row generally has worse performance than that of ``A'' row. A possible reason is that edges in graphs may have more uncertainty than attributes. Enforcing \emph{stability} on structures can help the model adapt to the uncertain edges in information aggregation and learn more robust node features.}
\textbf{2)}.Comparing perturbations on ``A'' and ``A+X'', DropEdge generally has better performance. Based on DropEdge, adding node attribute perturbations sometimes has positive effects while sometimes does not. \textcolor{black}{It is hard to find a proper distribution when adding perturbations on node attributes since the node attributes are usually heterogeneous. By contrast, DropEdge matches different graphs well and is more practical and easier for usage in practice.} 

\begin{figure*}[ht]
%第一行 loss
\centering
\begin{minipage}[t]{0.32\textwidth}
\centering
\includegraphics[width=\textwidth]{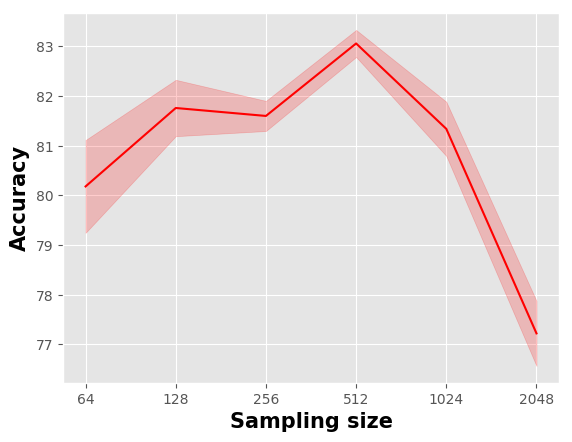}
% \vspace{-15pt}
\caption*{(a)}
\end{minipage}
\begin{minipage}[t]{0.32\textwidth}
\centering
\includegraphics[width=\textwidth]{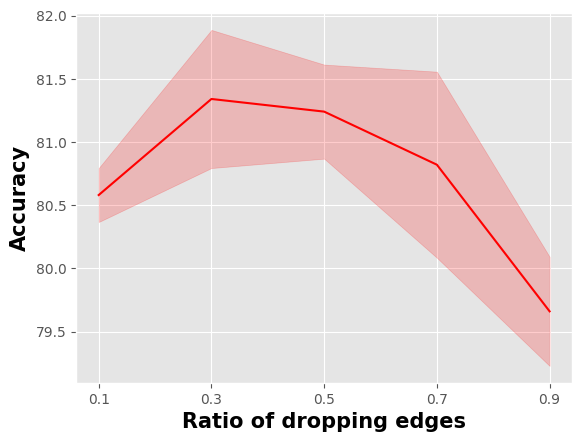}
% \vspace{-15pt}
\caption*{(b)}
\end{minipage}
\begin{minipage}[t]{0.32\textwidth}
\centering
\includegraphics[width=\textwidth]{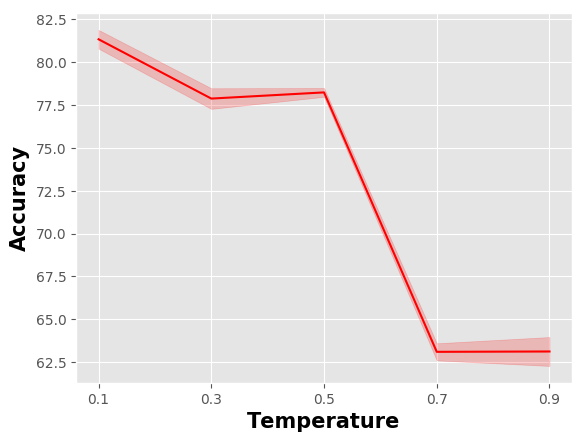}
% \vspace{-15pt}
\caption*{(c)}
\end{minipage} \\
\caption{The effect of different hyper-parameters on Pubmed. (a) indicates the sampling size $K$ . (b) indicates the ratio $\rho$ of dropping edges. (c) represents the temperature $\tau$ in the score function.}
\label{figure:hyperparams}
\end{figure*}
\subsubsection{\textcolor{black}{Hyper-parameter Analysis}}
In the proposed method, the ratio of dropping edges $\rho$ and the sampling size $K$ respectively control the variations of \emph{stability} and \emph{identifiability} of inputs. The temperature $\tau$ in the score function controls the similarity scale in contrastive learning. In order to explore the effects of these hyper-parameters, we further conduct an experiment to illustrate this. The results on Pubmed are shown in Figure~\ref{figure:hyperparams}. From this figure, we can summarize that: \textbf{1).} In Figure~\ref{figure:hyperparams} (a), 
% either a too small sampling size or a too large sampling size can lead to deteriorated performance. 
% As it is analysed in~\cite{tian2019contrastive}, NCE is a high biased and low variance estimator, and it requires a large sampling size $K$ to tighten the mutual information \emph{lower bound}. 
when $K$ becomes too large, there is a risk of sampling structure-similar nodes as negative pairs. In other words, too large $K$ can lead to inappropriate contrastive signals and deteriorates the quality of node representations. In general, the empirical $K=1024$ can provide satisfied performance. \textbf{2).} In Figure~\ref{figure:hyperparams} (b), a proper drop ratio $\rho$ tends to be around 0.5. \textcolor{black}{A too small $\rho$ leads to less variants of a node's neighbours. A too large $\rho$ will lose too much neighborhood information, which hinders the information propagation on graphs. In general, $\rho=0.3$ is an empirical value of dropedge technique~\cite{rong2019dropedge}.} \textbf{3).} The temperature $\tau$ controls the distribution of the score function. The method has its best performance when $\tau=0.1$. Empirically, $\tau \ll 0.1$ will cause value explosion in networks because of the exponential property. In summary, $\tau=0.1$ is an empirical value for satisfied model performance.
    % and is consistent with the choice in recent works~\cite{tian2019contrastive,hinton2012improving}.

\section{Conclusion and Future Work}
\textcolor{black}{In this paper, we propose a novel model SIGNNAP which illustrates the necessary properties of reliable node representations against perturbations on graphs.
% to learn reliable node representations against perturbations on graphs. 
Apart from the widely-used \emph{smoothness} property, SIGNNAP contains the \emph{stability} 
% where a node representation remains stable to slight perturbations on the input, 
and \emph{identifiability} property,
% where node representations of different structures are identifiable. SIGNNAP 
which provides a new insight of learning high-quality node representations for numerous graph algorithms. 
Through extensive experiments on various benchmarks, we show that SIGNNAP prevents the over-smoothing and over-fitting issue and learns more reliable node representations for downstream classification task.}

\textcolor{black}{However, there are still some limitations of our method. For example, the sampling method of negative pairs is random sampling, which would lead to the risk of sampling structure-similar representations of the anchor node. In fact, this case can be regarded as a positive unlabeled learning (PU learning)~\cite{kiryo2017positive} problem where we only have the positive and unlabeled data. Considering the unlabeled data as negative ones would cause the sampling bias and deteriorated performance.
In future, we would explore how to revise the negative sampling distribution by PU learning and correct the sampling bias.
}
% other sampling methods to better employ the contrastive objective for learning node embeddings against perturbations.

\section{Acknowledgements}
This work is supported by the National Key Research and Development Program of China (No. 2019YFB1804304), SHEITC (No. 2018-RGZN-02046), 111 plan (No. BP0719010), and STCSM (No. 18DZ2270700), and State Key Laboratory of UHD Video and Audio Production and Presentation. 
This work is also supported by ARC DP180100106 and DP200101328 of Australia. 
% Ivor W. Tsang and Yuangang Pan are also supported by A$^*$STAR Centre for Frontier AI Research (CFAR).

% \newpage
% \bibliographystyle{elsarticle-harv}      % basic style, author-year citations
% \bibliographystyle{spmpsci}      % mathematics and physical sciences
% \bibliographystyle{spphys}% APS-like style for physics
% \bibliography{reference}

% \bibliographystyle{abbrv}
\bibliographystyle{plainnat}
\bibliography{main}





% \end{thebibliography}

\clearpage
\appendix
\section{More Details about the Proofs}

\subsection{Proof of Lemma~\ref{lemma:lower_bound}}
\label{appendix:lower_bound}
In order to derive Lemma~\ref{lemma:lower_bound}, we first have the following Lemma for the score function $h_{\phi}$:
\begin{Lemma}
\label{lemma:critic_function}
The optimal value of score function $h_{\phi}(z_{0}^{1},z_{0}^{2})$ for minimizing loss $\mathcal{L}_{1}$ in Eq.~\ref{eq:contrastive_loss1} is proportional to the density ratio between the joint distribution $p(z_{0}^{1},z_{0}^{2})$ and the product of the marginals $p(z_{0}^{1})p(z_{0}^{2})$, which can be shown as follows:
\begin{equation}
\label{eq:optimal_critic}
h_{\phi}(z_{0}^{1},z_{0}^{2})\propto \frac{p(z_{0}^{1},z_{0}^{2})}{p(z_{0}^{1})p(z_{0}^{2})} 
% \propto \frac{p(z_{0}^{1}|z_{0}^{2})}{p(z_{0}^{1})}
\end{equation}
\end{Lemma}
where $\propto$ standards for 'proportional to'. The derivation regarding this lemma is provided in~\ref{appexdix:lemma}.

Considering Lemma~\ref{lemma:critic_function}, if we replace the score function in Eq.~\ref{eq:contrastive_loss1} with the density ratio in Eq.~\ref{eq:optimal_critic}, we have:
\begin{align}
    \mathcal{L}_{1}&= -\mathbb{E}_{\{z_{0}^{1},z_{0}^{2},z_{j}^{2}|_{j=1}^{K}\}} \Big [\log \frac{h_{\phi}(z_{0}^{1}, z_{0}^{2})}{\sum_{j=0}^{K}h_{\phi}(z_{0}^{1}, z_{j}^{2})} \Big ] \\
    &=-\mathbb{E}_{\{z_{0}^{1},z_{0}^{2},z_{j}^{2}|_{j=1}^{K}\}}\log \Big [ \frac{\frac{p(z_{0}^{1},z_{0}^{2})}{p(z_{0}^{1})p(z_{0}^{2})}}{\sum_{j=0}^{K}\frac{p(z_{0}^{1},z_{j}^{2})}{p(z_{0}^{1})p(z_{j}^{2})}} \Big ] \\
    &=\mathbb{E}_{\{z_{0}^{1},z_{0}^{2},z_{j}^{2}|_{j=1}^{K}\}}\log \Big [1+\frac{p(z_{0}^{1})p(z_{0}^{2})}{p(z_{0}^{1},z_{0}^{2})}\sum_{j=1}^{K}\frac{p(z_{0}^{1},z_{j}^{2})}{p(z_{0}^{1})p(z_{j}^{2})} \Big ] \\
    &=\mathbb{E}_{\{z_{0}^{1},z_{0}^{2}\}}\log \Big [1+\frac{p(z_{0}^{1})p(z_{0}^{2})}{p(z_{0}^{1},z_{0}^{2})}K\mathbb{E}_{z_{j}^{2}}\Big[ \frac{p(z_{0}^{1}|z_{j}^{2})}{p(z_{0}^{1})} \Big]
    \Big ] \\
    &= \mathbb{E}_{\{z_{0}^{1},z_{0}^{2}\}}\log \Big [1+\frac{p(z_{0}^{1})p(z_{0}^{2})}{p(z_{0}^{1},z_{0}^{2})}K \Big ] \\
    &\geq \log(K)-\mathbb{E}_{\{z_{0}^{1},z_{0}^{2}\}}\log \Big [ \frac{p(z_{0}^{1},z_{0}^{2})}{p(z_{0}^{1})p(z_{0}^{2})} \Big ] \\
    &=\log(K) - \mathcal{I}(z_{0}^{1},z_{0}^{2})
\end{align}
Similarly for the loss function $\mathcal{L}_{2}$ in Eq.~\ref{eq:contrastive_loss2}, we have:
\begin{align}
    \mathcal{L}_{2}\geq \log(K) - \mathcal{I}(z_{0}^{1},z_{0}^{2})
\end{align}
By combining the derivation together, we have the following formulation:
\begin{align}
\label{eq:final_MI_ELBO}
    \mathcal{I}(z_{0}^{1},z_{0}^{2})\geq \log(K) - \frac{1}{2}(\mathcal{L}_{1}+\mathcal{L}_{2})\geq \log(K) - \mathcal{L}
\end{align}
From Eq.~\ref{eq:final_MI_ELBO}, we can see that when minimizing our contrastive objective function $\mathcal{L}$ in Eq.~\ref{eq:objective_function}, the \emph{lower bound} of the mutual information $\mathcal{I}(z_{0}^{1},z_{0}^{2})$ is maximized.

\subsection{Proof of Lemma~\ref{lemma:critic_function}}
\label{appexdix:lemma}
The derivation of Lemma~\ref{lemma:critic_function} is derived as follows.
The objective function in Eq.~\ref{eq:contrastive_loss1} is indeed a categorical cross-entropy that classifies the positive sample correctly. To clarify this lemma,
for $z_{0}^{1}$, we denote $z_{0}^{2}$ as a positive sample and $z_{j}^{2}$ as a negative sample.
Let us write the optimal probability for minimizing $\mathcal{L}_{1}$ in Eq.~\ref{eq:contrastive_loss1} as $p(d=0|z_{0}^{1},z_{0}^{2},z_{j}^{2}|_{j=1}^{K})$ where $[d=0]$ being the indicator that $z_{0}^{2}$ is the positive sample. Then we can have:
\begin{align}
p(d=0|z_{0}^{1},z_{0}^{2},z_{j}^{2}|_{j=1}^{K})&=\frac{p(z_{0}^{2}|z_{0}^{1})\prod_{i\neq 0}p(z_{i}^{2})}{\sum_{j=0}^{K}p(z_{j}^{2}|z_{0}^{1})\prod_{i\neq j}p(z_{i}^{2})} \\
&=\frac{\frac{p(z_{0}^{2}|z_{0}^{1})}{p(z_{0}^{2})}}{\sum_{j=0}^{K}\frac{p(z_{j}^{2}|z_{0}^{1})}{p(z_{j}^{2})}} \\
&=\frac{\frac{p(z_{0}^{1},z_{0}^{2})}{p(z_{0}^{1})p(z_{0}^{2})}}{\sum_{j=0}^{K}\frac{p(z_{0}^{1},z_{j}^{2})}{p(z_{0}^{1})p(z_{j}^{2})}}
\end{align}
By comparing the above equation with $\mathcal{L}_{1}$, we can see that the optimal value of $h_{\theta}(z_{0}^{1},z_{0}^{2})$ in Eq.~\ref{eq:contrastive_loss1} is proportional to $\frac{p(z_{0}^{1},z_{0}^{2})}{p(z_{0}^{1})p(z_{0}^{2})}$.

\begin{table*}[]
\centering
\caption{The hyper-parameter setting of different models on Pubmed, Facebook, Coauthor-CS and Coauthor-Phy. ``-'' indicates the number of layers is the same as the backbone model. ``/'' means the method does not have this parameter. On Amazon-Com and Amazon-Pho, the only difference is the weight decay is set as 0.0 for all methods.}
\label{table:hyper_params_setting}
\renewcommand{\arraystretch}{1.2}
\setlength{\tabcolsep}{1.0mm}{ 
\scalebox{0.8}{
\begin{tabular}{l|l|l}
\hline
Model     & nlayers & Hyper-parameters                                                                                                                                                                                                               \\ \hline
GCN       & 2       & lr:0.003, weigt\_decay:5e-3, dropout:0.5, n\_hidden:128, iterations:1,000                                                                                                                                                      \\ \hline
ResGCN    & 3       & lr:0.003, weight\_decay:5e-3, dropout:0.5, n\_hidden:128, iterations:1,000                                                                                                                                                      \\ \hline 
JKNet     & 3       & lr:0.003, weight\_decay:5e-3, dropout:0.5, n\_hidden:128, iterations:1,000                                                                                                                                                      \\ \hline
GraphSage & 2       & lr:0.01, weight\_decay:5e-4, dropout:0.5, n\_hidden:128, iterations:1,000                                                                                                                                                       \\ \hline
GAT       & 2       & lr:0.005, weight\_decay:5e-4, num\_heads:8, num\_out\_heads=8,  iterations:1,000                                                                                                                                                \\ \hline
DropEdge  & -       & \begin{tabular}[c]{@{}l@{}}The hyper-parameters follow the settings of different backbones. \\ The drop edge ratio is 0.3.\end{tabular}                                                                                 \\ \hline
DeepWalk       & /       & \begin{tabular}[c]{@{}l@{}}num\_walks:10, walk\_length:80, window\_size:10, n\_hidden:128, iterations:200. \end{tabular}           \\
\hline
Node2Vec       & /       & \begin{tabular}[c]{@{}l@{}}p:1.0, q:0.5, num\_walks:10, walk\_length:80, window\_size:10, \\ 
n\_hidden:128, iterations:200. \end{tabular}           \\
\hline
ARWMF       & /       & \begin{tabular}[c]{@{}l@{}}rank\_K:200, window\_size:5, lr:0.001, iterations:1000. \end{tabular}           \\
\hline
DGI       & 2       & \begin{tabular}[c]{@{}l@{}}lr:0.001, weight\_decay:0.0, dropout:0.0, n\_hidden:512, iterations:300. \end{tabular}           \\
\hline
CMV       & 2       & \begin{tabular}[c]{@{}l@{}}lr:0.001, sampling\_size:1000, dropout:0.5, n\_hidden:512, iterations:3000. \end{tabular}           \\
\hline
GCC       & 2       & \begin{tabular}[c]{@{}l@{}}lr:0.005, subgraph\_size:128, sampling\_size:32,temperature:0.07,\\ dropout:0.5, n\_hidden:64, iterations:1000. \end{tabular}           \\
\hline
SIGNNAP    & -       & \begin{tabular}[c]{@{}l@{}}lr:0.001, The other hyper-parameters follow the settings of different backbones. \\ The drop edge ratio $\rho$ is 0.3, the temperature $\tau$ is 0.1 \\ and the sampling size $K$ is 1024.\end{tabular} \\ \hline
\end{tabular}
}}
\end{table*}
\section{More Details about the Experiments}
\label{appen:more_experiments}
\subsection{Detailed Parameter Settings of Different Models}
\label{sec:params_setting}
More detailed parameter settings are shown in Table~\ref{table:hyper_params_setting}. 
Note that for some datasets such as Pubmed and Coauthor-CS which have been reported in many works, the hyper-parameter settings are the same with the original paper or codes\cite{kipf2017semi,wang2019kgat,velickovic2018deep,shchur2018pitfalls}. For Facebook, the above hyper-parameter settings also perform well and we directly use them for Facebook.

\section{More Experiments}

\begin{figure*}[ht]
%第一行 loss
\centering
\begin{minipage}[t]{0.32\textwidth}
\centering
\includegraphics[width=\textwidth]{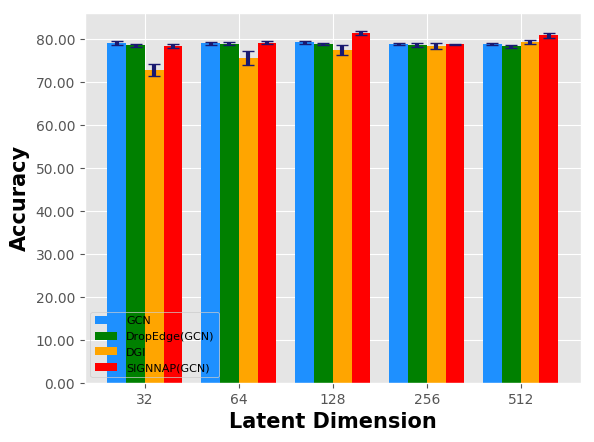}
% \vspace{-15pt}
\caption*{(a) Pubmed}
\end{minipage}
\begin{minipage}[t]{0.32\textwidth}
\centering
\includegraphics[width=\textwidth]{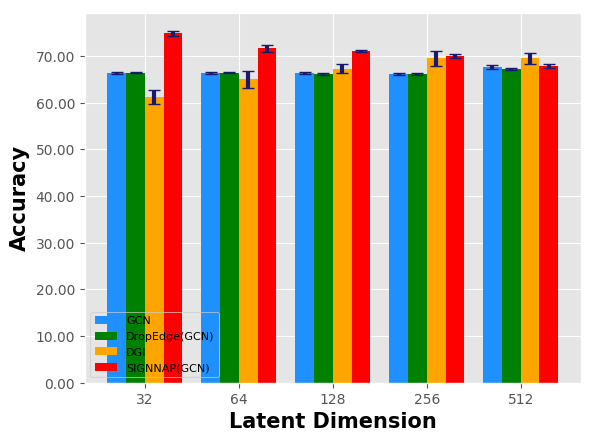}
% \vspace{-15pt}
\caption*{(b) Facebook}
\end{minipage}
\begin{minipage}[t]{0.32\textwidth}
\centering
\includegraphics[width=\textwidth]{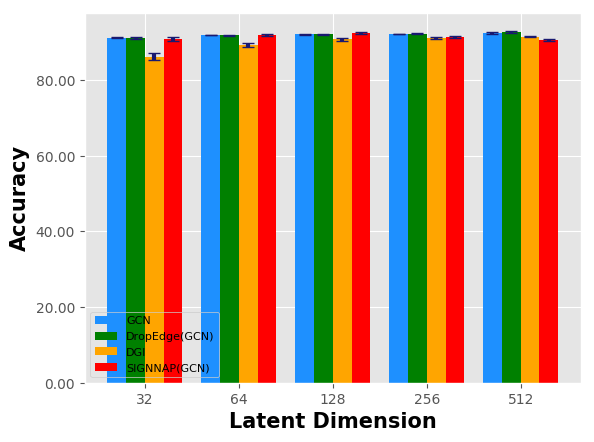}
% \vspace{-15pt}
\caption*{(c) Coauthor-CS}
\end{minipage} \\
\caption{The effect of different latent dimensions on different benchmarks. For clarity, we show the performance of some representative methods such as GCN, DropEdge(GCN), DGI and SIGNNAP(GCN). The standard deviation of each bar is also shown in this figure.}
\label{figure:different_d}
\end{figure*}
\subsection{Different Latent Dimensions}
\label{appen:different_d}
For representation learning on graphs, the latent dimension is also an important factor that influences the model performance. We conduct an experiment to more comprehensively compare the performance of different methods. The results are shown in Figure~\ref{figure:different_d}. From this figure, we have the following observations: \textbf{1).} Compared to the supervised methods (i.e. GCN and DropEdge(GCN)) on different dimensions, the proposed unsupervised method SIGNNAP(GCN) reaches competitive performance and even exceeds the supervised ones on different benchmarks. \textbf{2).} The performance of DGI varies a lot along different dimensions and has its best performance when the latent dimension is 512. By contrast, SIGNNAP(GCN) provides more stable and better performance along different latent dimensions.
\vspace{-8pt}
\begin{itemize}
    \item Compared to the supervised methods (i.e. GCN and DropEdge(GCN)) on different dimensions, the proposed unsupervised method SIGNNAP(GCN) reaches competitive performance and even exceeds the supervised ones on different benchmarks.
    % \vspace{-5pt}
    \item The performance of DGI varies a lot along different dimensions and has its best performance when the latent dimension is 512. By contrast, SIGNNAP(GCN) provides more stable and better performance along different latent dimensions.
\end{itemize}
\vspace{-8pt}

\end{document}